\documentclass{article}

\usepackage[utf8]{inputenc} % allow utf-8 input
\usepackage[T1]{fontenc}    % use 8-bit T1 fonts
\usepackage{hyperref}       % hyperlinks
\usepackage{url}            % simple URL typesetting
\usepackage{booktabs}       % professional-quality tables
\usepackage{amsfonts}       % blackboard math symbols
\usepackage{nicefrac}       % compact symbols for 1/2, etc.
\usepackage{microtype}      % microtypography
\usepackage{xcolor}         % colors
\usepackage{graphicx}
\usepackage{amsmath}
\usepackage{amsfonts}
\usepackage{subcaption}
\usepackage{caption}
\usepackage{float}
\usepackage{mathtools}
\usepackage{extpfeil} 
\usepackage{natbib}
\usepackage{wrapfig}
\usepackage{overpic}
\usepackage{xspace}
\usepackage{enumitem}
\usepackage{xurl}

\usepackage{multirow}
\usepackage{multicol}
\usepackage{siunitx}
\usepackage{bm}
\usepackage{listings}
\usepackage{tikz}
\usepackage{tabularx}   % 使得 tabularx 环境可用
\usepackage{tabularray} % 更现代的表格控制包

\usepackage{authblk}

\lstdefinelanguage{C++}{
  language=C,
  morekeywords={
    class, public, private, protected, template, typename,
    std, cout, cin, endl, string, vector, map, unordered_map,
    auto, constexpr, noexcept, nullptr, using, namespace, new, delete
  },
  sensitive=true,
  morecomment=[l]{//},
  morecomment=[s]{/*}{*/},
  morestring=[b]",
}

\lstdefinelanguage{PDDL}{
    keywords={
        :init, :goal, and, not, or, implies, when,
        on_rack, in_pot, can_wipe_table, in_drawer,
        hand_free, on, in_bowl, wiped, is_open,
        remove_lid, pick_from_rack, pick_from_pot,
        place_on_table, put_in_bowl, wipe_table,
        open_drawer, pick_from_drawer, place_in_drawer,
        close_drawer
    },
    keywordstyle=\color{blue}, % 与 C++ 关键字风格一致
    sensitive=false,
    morecomment=[l];,          % 单行注释以 ; 开头
    morestring=[b]"            % 字符串用双引号
}

\usepackage[final]{neurips_2025}

\newcommand{\algname}{UniDomain\xspace}

\newcommand{\alldomains}{unified domain\xspace} % The complete knowledge base containing all minimal domains learned from demonstrations
\newcommand{\Alldomains}{Unified Domain\xspace} % The complete knowledge base containing all minimal domains learned from demonstrations

\newcommand{\fuseddomain}{meta-domain\xspace} % A unified PDDL domain obtained by fusing a relevant subset of domains
\newcommand{\fuseddomains}{meta-domains\xspace} % A unified PDDL domain obtained by fusing a relevant subset of domains

\newcommand{\singledomain}{atomic domain\xspace} % A minimal PDDL domain derived from a single demonstration
\newcommand{\singledomains}{atomic domains\xspace} % A minimal PDDL domain derived from a single demonstration

\newcommand{\knowledgegraph}{knowledge graph\xspace} % The fused meta-domain considered as a knowledge graph for a task group.

\newcommand{\taskspace}{task space\xspace} % The entire set of robotic manipulation tasks considered
\newcommand{\taskgroup}{task class\xspace} % A group of related manipulation tasks sharing common features, solvable using one meta-domain
\newcommand{\tasks}{task instances\xspace} % A single, specific robotic manipulation task to be solved

\newcommand{\image}{\ensuremath{I}\xspace} % image of the initial scene
\newcommand{\instruct}{\ensuremath{T}\xspace} % language instruction of the task

\newcommand{\domain}{\ensuremath{D}\xspace}  % a PDDL domain
\newcommand{\problem}{\ensuremath{P}\xspace} % a PDDL problem

\newcommand{\oprset}{\ensuremath{\mathcal{O}}\xspace}  % operator set
\newcommand{\prset}{\ensuremath{\mathcal{P}}\xspace}  % predicate set
\newcommand{\negprset}{\ensuremath{\lnot\mathcal{P}}\xspace} % negative predicate set
\newcommand{\opr}{\ensuremath{o}\xspace}  % a single operator
\newcommand{\pre}{\ensuremath{p}\xspace}  % a single predicate
\newcommand{\goal}{\ensuremath{s_g}\xspace} % goal state
\newcommand{\init}{\ensuremath{s_0}\xspace} % initial state
\newcommand{\objset}{\ensuremath{\mathcal{B}}\xspace}    % objects (renamed to \mathcal{B} to avoid collision)

\newcommand{\actseq}{\ensuremath{A}\xspace} % the solution, plan, represented as action sequence

\title{\algname: Pretraining domains from VLA datasets for Vision-Language Task Planning of Robots}
\title{\algname: Pretraining a Unified PDDL Domain \\from Real-World Demonstrations\\ for Generalizable Robot Task Planning}

\author{
\textbf{Haoming Ye}\textsuperscript{1,2 *},
\textbf{Yunxiao Xiao}\textsuperscript{2,3 *},
\textbf{Cewu Lu}\textsuperscript{1,2},
\textbf{Panpan Cai}\textsuperscript{1,2 \dag} \\
\textsuperscript{1}Shanghai Jiao Tong University \quad
\textsuperscript{2}Shanghai Innovation Institute \\
\textsuperscript{3}Beijing University of Posts and Telecommunications
}
\begin{document}

\maketitle
\vspace{-2.8em}
{\centering
\footnotesize
* Equal contribution \\
\dag\ Corresponding author: \texttt{cai\_panpan@sjtu.edu.cn}\par
}

% {\renewcommand{\thefootnote}{*}
%  \footnotetext{These authors contributed equally to this work.}
% }
% {\renewcommand{\thefootnote}{\dag}
%  \footnotetext{Corresponding author: \texttt{cai\_panpan@sjtu.edu.cn}}
% }

% \vspace{-3.5em}
% % --------- Author notes + acceptance + logos (inline under authors) ----------
% %（如果与标题间距仍偏大，可在这行上方再加：\vspace{-1.0em}）
% \begin{center}
%   \begin{minipage}{0.92\linewidth}\centering
%     \small
%     \textsuperscript{*} These authors contributed equally to this work.\quad
%     \textsuperscript{\dag} Corresponding author
%   \end{minipage}

%   \vspace{0.3em} % 原 0.5em -> 更紧
%   {\bfseries Accepted at NeurIPS 2025}

%   \vspace{0.35em} % 原 0.6em -> 更紧
%   % 建议尽量使用矢量(PDF)或高分辨率图片；高度可调到 1.6–2.0cm
%   \raisebox{-0.5\height}{\includegraphics[height=2.5cm]{images/sjtu.jpg}}\hspace{1.2em}%
%   \raisebox{-0.5\height}{\includegraphics[height=2.5cm]{images/sii.jpg}}
% \end{center}

\begin{abstract}
Robotic task planning in real-world environments requires reasoning over implicit constraints from language and vision. While LLMs and VLMs offer strong priors, they struggle with long-horizon structure and symbolic grounding. Existing methods that combine LLMs with symbolic planning often rely on handcrafted or narrow domains, limiting generalization. We propose \algname, a framework that pre-trains a PDDL domain from robot manipulation demonstrations and applies it to online robotic task planning. It extracts \singledomains from {12,393} manipulation videos to form a \alldomains with {3,137} operators, {2,875} predicates, and {16,481} causal edges. Given a target class of tasks, it retrieves relevant atomics from the \alldomains and systematically fuses them into high-quality \fuseddomains to support compositional generalization in planning. 
Experiments on diverse real-world tasks show that \algname solves complex, unseen tasks in a zero-shot manner, achieving up to 58\% higher task success and 160\% improvement in plan optimality over state-of-the-art LLM and LLM-PDDL baselines. \footnote{The code and demonstration video are available at: \url{https://roboticsjtu.github.io/UniDomain/}}
% On 100 unseen real-world tasks, \algname outperforms state-of-the-art LLM and LLM-PDDL baselines by up to 58\% in success rate and 160\% in plan optimality.
\end{abstract}

% \footnotetext[1]{The code and demonstration video are available at: \url{https://unidomain.github.io/}}

\section{Introduction}

Robotic task planning in real-world environments requires reasoning over complex constraints that are often implicitly specified in natural language instructions and grounded in visual observations. For instance, the task \textit{``partition the stack into even and odd numbers, sorted in ascending order''} implicitly encodes long-term dependencies involving unstacking, sorting, and placement. Similarly, \textit{``make a cup of tea''} entails a sequence of preparatory steps such as opening the cabinet, finding the tea cup, and boiling the water. These tasks demand structured reasoning over action preconditions, temporal dependencies, and physical affordances, in order to ensure safety (e.g., avoiding spills), prevent irreversible states, and minimize human intervention.

However, these problems remain fundamentally challenging: instructions are open-ended, scenes are unstructured, and constraints are implicit. Recent approaches leverage the commonsense priors of Large Language Models (LLMs) and Vision-Language Models (VLMs) \cite{gpt-4o, qwen2.5VL} to generalize across real-world tasks. Yet, despite their strengths in language and visual understanding, LLMs and VLMs often fail to model action preconditions and effects accurately, and struggle with generating coherent, long-horizon plans \cite{onthelimit}.
To improve reasoning, recent work~\cite{llm+p, llm+dp, pddlego} integrates LLMs with symbolic planning using the Planning Domain Definition Language (PDDL)~\cite{pddl}. A common pipeline translates language instructions and scene images into structured PDDL domain and problem files, then invokes a symbolic planner~\cite{fastdownward} to produce a plan. While LLMs can reliably generate PDDL problems when the domain is given~\cite{autogpt+p, ViLaIn}, they struggle to construct realistic domain files \cite{llm+dm, interpret} for planning tasks. Most models are only familiar with abstract domains \cite{nl2plan, BoN-iVML}  such as \texttt{BlocksWorld}, \texttt{Logistics}, and \texttt{Rovers}, and lack grounding in real-world robot interactions.

\begin{figure}[!t]
    \centering
    \includegraphics[width=1\linewidth]{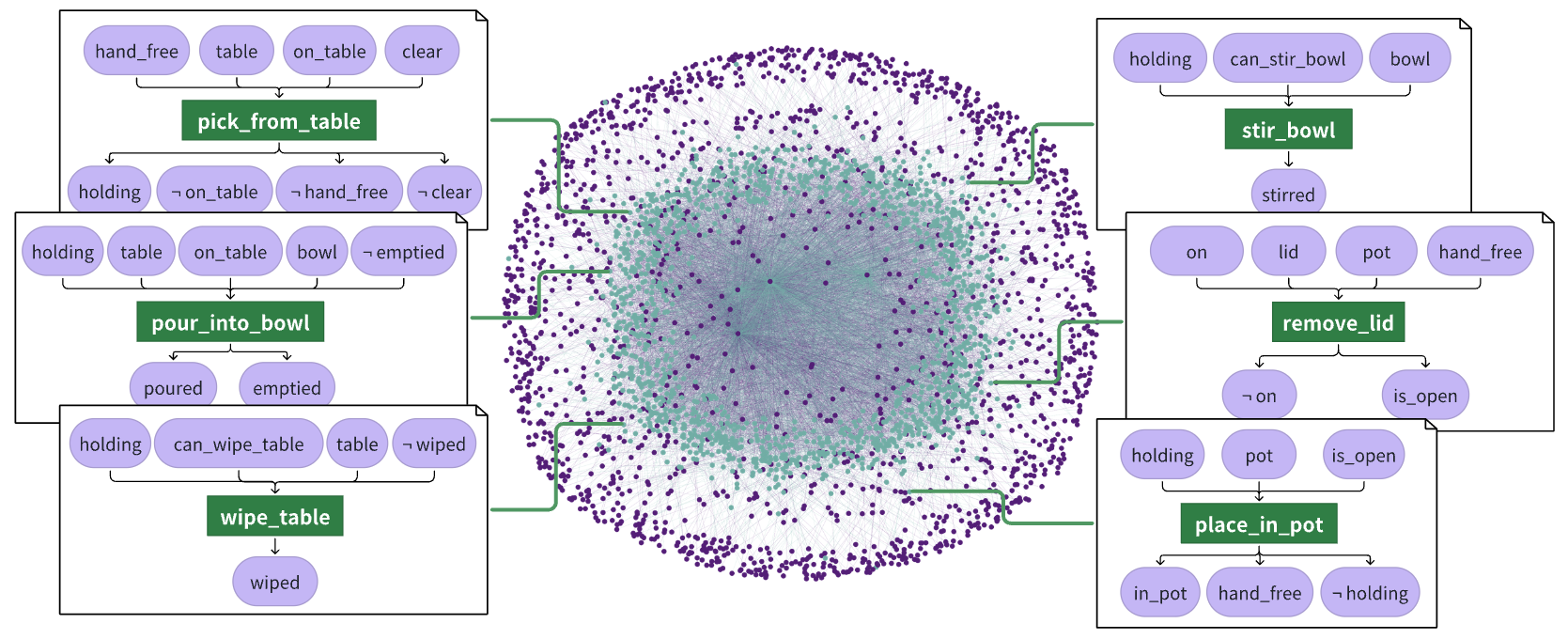}
    \caption{Visualization of our pre-trained \alldomains, with 3,137 operator nodes (green) and 2,875 predicate nodes (purple).}
    \label{fig:Knowledge Graph}
    \vspace*{-0.2cm}

\end{figure}

To overcome this limitation, we propose leveraging large-scale demonstration datasets designed for training Vision-Language-Action (VLA) models \cite{rdt, openvla, pi0} (e.g., DROID~\cite{droid}). These datasets are grounded in real robot executions. Visual information in demonstrations captures the actual preconditions and effects of robot actions in diverse environments. Although each demonstration typically covers a single operation, the collection spans a broad spectrum of manipulation tasks, enabling \textit{compositional generalization}. By learning \singledomains from \textbf{12,393} real-world demonstrations and merging them into a structured \alldomains, we construct a connected symbolic \knowledgegraph (Figure \ref{fig:Knowledge Graph})---containing \textbf{3,137} operators, \textbf{2,875} predicates, and \textbf{16,481} causal edges---to support long-term planning under complex constraints across diverse household tasks.

Particularly, we introduce \algname, a framework for \emph{pre-training a unified, general-purpose PDDL domain} (referred to as the \textit{\alldomains}) from large-scale robot manipulation demonstrations, and applying it for online planning in unseen tasks. \algname comprises three stages:
% \begin{itemize}
(1) \textit{Domain Pretraining.} Given a demonstration dataset, each video is segmented into keyframes using an energy-based method. A VLM proposes an initial \singledomain from the keyframes, which is refined via closed-loop verification with an LLM to ensure syntactic correctness, solvability, and commonsense alignment. The resulting set of \singledomains constitutes a large-scale \alldomains.
(2) \textit{Domain Fusion.} For any targeted \taskgroup, a relevant subset of \singledomains is retrieved from the large-scale \alldomains and systematically fused into a high-quality \textit{\fuseddomain}. The fusion merges functionally overlapping predicates and operators, yielding a compact yet expressive subgraph for generalizable task planning.
(3) \textit{Online Planning.} Given a specific task, \algname constructs a grounded PDDL problem with the high-quality \fuseddomain and solves it using a PDDL planner to generate an optimal plan.
% \end{itemize}

We analogize these stages to the \textit{pre-training}, \textit{post-training}, and \textit{inference} phases of foundation models \cite{pretraining, posttraining}. \emph{Domain pretraining} builds a comprehensive \alldomains encoding general manipulation knowledge. \emph{Domain fusion} constructs a specialized \fuseddomain with enhanced symbolic connectivity. \emph{Online planning} applies the \fuseddomain in a zero-shot fashion, to solve unseen tasks without additional demonstrations or feedback.
% The learned \alldomains spans \textbf{3137} operators, \textbf{2875} predicates, and \textbf{16481} causal edges. 
% An example \fuseddomain used in our experiments contains 61 actions, 78 predicates, and 354 causal edges.

We evaluate \algname on four real-world task domains unseen during training, comprising 100 long-horizon tasks with complex constraints. Results show that \algname significantly outperforms popular LLM-only planners (e.g., Code-as-Policies \cite{cap}, ReAct \cite{react}) and state-of-the-art hybrid LLM-PDDL baselines (e.g., ISR-LLM \cite{isr-llm}, BoN-iVML \cite{BoN-iVML}), achieving up to 58\% higher task success and 160\% better plan optimality than the strongest baselines. Ablation studies confirm that performance gains stem from data-driven domain learning, closed-loop verification, hierarchical fusion for \fuseddomain construction, and task-relevant grounding during online planning. 

In summary, the main contributions of this work include:
% \begin{itemize}
    (1) The first framework to pre-train a unified PDDL domain for high-level robot task planning from large-scale, real-world demonstrations;
    (2) A novel LLM-based domain fusion method for combining small, disconnected PDDL domains into a coherent and compact \fuseddomain thus supporting compositional generalization;
    and (3) A novel online task planner that applies the fused \fuseddomain to solve general, unseen tasks through VLM-grounded PDDL planning.

\section{Background and Related Work}
\subsection{PDDL Fundamentals}
The Planning Domain Definition Language (PDDL) \cite{pddl} formalizes classical planning problems as a tuple $(\domain, \problem)$, where $\domain$ is a \textit{domain} and $\problem$ is a \textit{problem instance}. The domain is defined as $\domain = (\oprset, \prset)$, consisting of a set of operators $\oprset$ and predicates $\prset$. Each predicate $\pre \in \prset$ is a Boolean-valued function over typed objects, representing properties or relations among entities. Each operator $\opr \in \oprset$ is defined by its preconditions $\text{pre}(\opr) \subseteq \prset \cup \negprset$ and effects $\text{eff}(\opr) \subseteq \prset \cup \negprset$, where $\negprset$ denotes the set of negated predicates. These preconditions and effects describe the logical requirements and state transitions induced by executing $\opr$.
% The Planning Domain Definition Language (PDDL) formalizes classical planning problems as a tuple $(\domain, \problem)$, where \domain is a \textit{domain} and \problem is a \textit{problem instance}. The domain is defined as $\domain = (\oprset, \prset)$, consisting of an operator set \oprset and predicate set \prset. Each predicate $\pre \in \prset$ is a Boolean-valued function over typed objects, representing properties or relations among entities. Each operator $\opr \in \oprset$ is defined with preconditions $\text{pre}(\opr) \subseteq \prset \cup \negprset$ and effects $\text{eff}(\opr) \subseteq \prset \cup \negprset$, \textcolor{red}{where \negprset denotes the set consisted of Logical NOT of all predicates in \prset}, describing the logical requirements and state transitions induced by executing $\opr$.
The problem instance is defined as $\problem = (\objset, \init, \goal)$, where \objset is the object set, \init is the initial state given as a grounded conjunction of predicates, and \goal is a partially specified target state. A planner searches for an action sequence $\actseq = \langle a_1, a_2, \ldots, a_T \rangle$, where each $a_t$ is a grounded instance of an operator from \oprset, such that applying $\actseq$ to \init results in a state satisfying \goal.
Given $(\domain, \problem)$, off-the-shelf symbolic planners such as Fast Downward~\cite{fastdownward} use heuristic search to compute a valid plan that transitions the world from \init to a goal state. The symbolic structure of PDDL enables interpretable, verifiable, and constraint-aware planning in complex domains.

\subsection{LLM-based Task Planning}
Large Language Models (LLMs) \cite{gpt-4o, deepseekv3} exhibit strong commonsense reasoning and structured generation, making them appealing for robotic task planning. Recent work translates free-form instructions into code-like plans (e.g., Code-as-Policies~\cite{cap}, ProgPrompt~\cite{progprompt}), filters actions using affordance and cost (SayCan~\cite{saycan}, SayCanPay~\cite{saycanpay}), or integrates feedback via closed-loop reasoning (ReAct~\cite{react}, Inner Monologue~\cite{innermonologue}, Reflexion~\cite{reflexion}). Others incorporate search (LLM-MCTS~\cite{llm-mcts}), symbolic grounding (Chain-of-Symbol~\cite{cos}), or multimodal inputs (ViLa~\cite{vila}). Yet, LLMs alone often fail to enforce implicit constraints and produce coherent long-horizon plans, motivating their integration with structured symbolic systems, such as PDDL.

\subsection{Integration of PDDL and LLM Planning}

Recent work has explored hybrid LLM-PDDL approaches to task planning. One common paradigm fixes the PDDL domain and uses LLMs to generate problems. For example, LLM+P~\cite{llm+p}, AutoGPT+P~\cite{autogpt+p}, and ViLaIn~\cite{ViLaIn} translate natural language into PDDL problems conditioned on predefined domains, then solve them using classical planners. Some frameworks~\cite{llm+dp, pddlego} combine this with adaptive planning. However, their reliance on manually crafted domains requires significant effort from PDDL experts and restricts generalization across tasks and environments.

Several works attempted to construct PDDL domains directly from language instructions. ISR-LLM~\cite{isr-llm} generates a domain-problem pair from language. NL2Plan~\cite{nl2plan} uses five stages of LLM-based verification to iteratively construct a PDDL domain. BoN-iVML~\cite{BoN-iVML} uses Best-of-$N$ sampling followed by iterative refinement to generate a domain. The domains generated by these methods have limited quality due to restricted information.
Other work attempts to iteratively construct PDDL domains through external feedback. InterPreT~\cite{interpret} and LLM-DM~\cite{llm+dm} incorporate human-in-the-loop to iteratively refine operators. Ada~\cite{ada} and LASP~\cite{lasp} use environment feedback to refine the domain. While effective in simulation, these methods are impractical for scalable deployment due to the cost, latency, and noise in external feedback. In contrast, our method uses internal closed-loop validation with synthetic test problems, which automates domain learning without human input.

There also exist a few works that learn PDDL domains from visual demonstration. BLADE~\cite{blade} learns a domain from a robot manipulation trajectory with a given set of actions. pix2pred~\cite{pix2pred} extracts operators from visual demonstrations given a predefined set of predicates. Diehl et al. \cite{2021iros} and Huang et al. \cite{2025icra} automate the generation of a robotic planning domain from single or a few repeated demonstrations in simulation. These works typically aim to generate a narrow domain tied to the demonstrated task, and require task-specific priors (action, predicate sets or in-domain demonstrations) as input.
\algname, however, learns a unified, general-purpose PDDL domain from large-scale robotic manipulation datasets, to enable compositional generalization and support symbolic task planning in diverse manipulation tasks under complex constraints.

\section{Method Overview} \label{sec:overview}

\begin{figure}[!t]
    \centering
    \includegraphics[width=\linewidth]{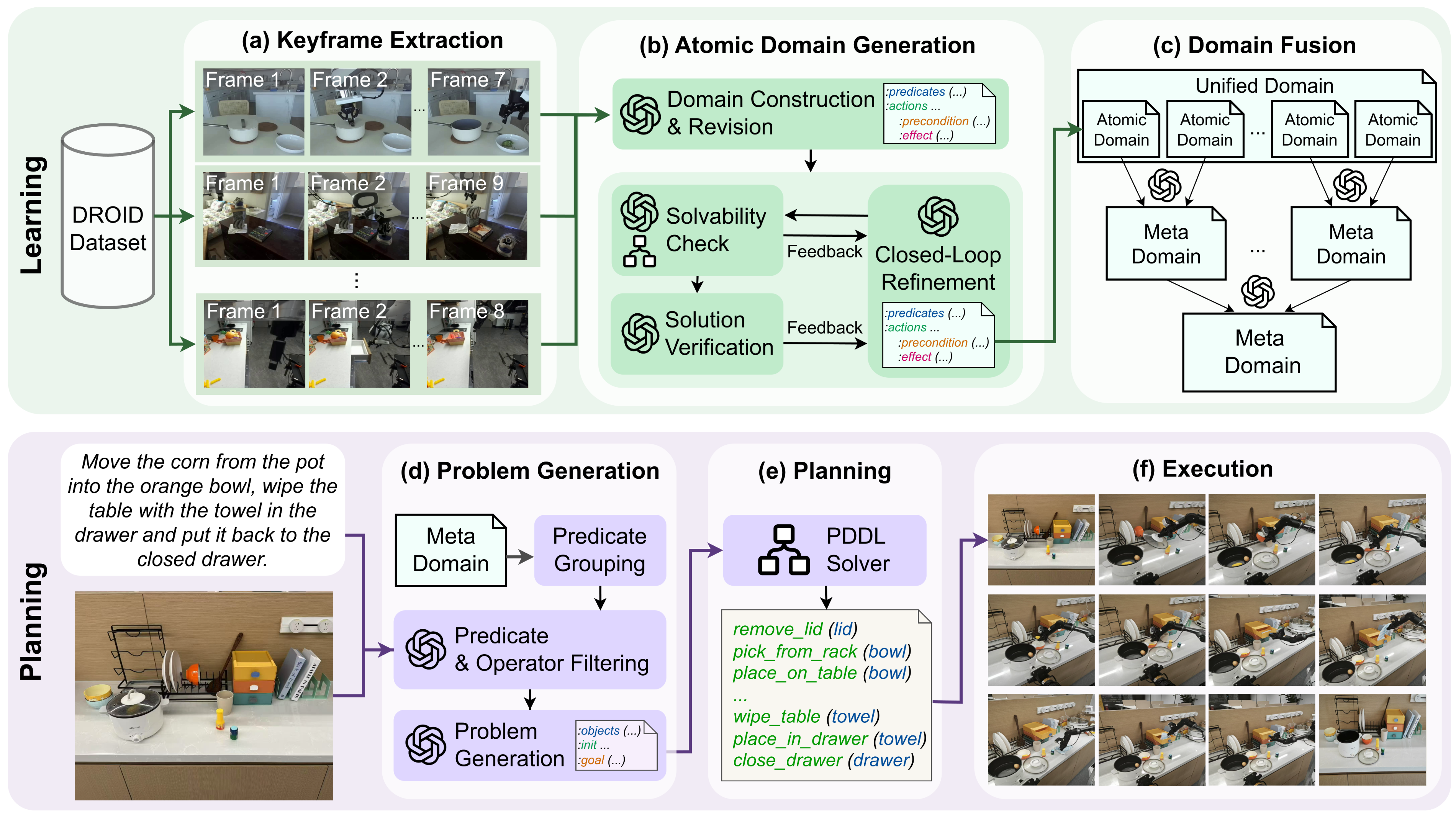}
    \caption{Overview of \algname. See detailed descriptions in Section \ref{sec:overview}.}
% From visual demonstrations, \algname (a) extracts keyframes, (b) generates and verifies task-specific PDDL domains, and (c) fuses them into a unified domain. At test time, it (d) grounds a task instruction and scene into a \textcolor{red}{PDDL problem with filtered domain}, (e) plans with a symbolic solver, and (f) executes the resulting action sequence.}
    \label{fig:overview}
    \vspace*{0.1cm}

\end{figure}

\textit{\algname} (Figure~\ref{fig:overview}) is a three-phase framework for vision-language-conditioned task planning with symbolic structure. In the first phase, \textit{Domain Pretraining}, atomic PDDL domains are extracted from visual-language robot demonstrations using keyframe extraction (Figure~\ref{fig:overview}a), VLM-based domain construction, and LLM-based closed-loop refinement (Figure~\ref{fig:overview}b). These domains collectively form a \alldomains capturing broad manipulation knowledge. In the second phase, \textit{Domain Fusion} (Figure~\ref{fig:overview}c), task-class-relevant \singledomains are retrieved and hierarchically merged into a compact \fuseddomain by aligning functionally-overlapping predicates and operators. In the final phase, \textit{Online Planning}, a task instruction and scene image are used to construct a grounded PDDL problem (Figure~\ref{fig:overview}d), which is solved with a classical planner (Figure~\ref{fig:overview}e) using the fused meta-domain to generate executable plans (Figure~\ref{fig:overview}f).

\section{Domain Pretraining}
This section pre-trains a general PDDL domain for robot manipulation tasks, which aggregates a large set of \singledomains learned from the DROID~\cite{droid} dataset.
Given each visual-language demonstration, \algname generates an \singledomain via two steps: unsupervised keyframe segmentation (Section~\ref{sec:key_frame}) and LLM-guided domain construction with closed-loop verification (Section~\ref{sec:domain_generation}). 

\subsection{Energy-Based Keyframe Extraction} \label{sec:key_frame}
Keyframe extraction is a well-studied problem, though many robotics approaches rely on task-specific signals such as human-object contact or end-effector trajectories~\cite{dexmimicgen, yoto}. Other methods~\cite{similarity} extract keyframes based on frame similarity using embeddings from pretrained vision-language models like CLIP~\cite{clip} or SigLIP~\cite{siglip}, but these approaches incur high computational costs. We propose a simple, domain-agnostic method that identifies semantic transitions in videos by detecting changes in grayscale intensity (Figure~\ref{fig:overview}a). Let $\image_t \in \mathbb{R}^{W \times H}$ denote the grayscale version of the input frame $t$. We define the frame energy as 

\begin{equation}
    E(\image_t) = \sum_{i=1}^W \sum_{j=1}^H \image_t(i, j)^2.
\end{equation}

The energy sequence ${E(\image_t)}_{t=1:T}$ is computed across all frames, and keyframes are selected by identifying local extrema using a sliding window of size $K$: a frame $t$ is chosen as a keyframe if
\begin{equation}
    E_t = \max_{i \in [t-K, t+K]} E_i \quad \text{or} \quad E_t = \min_{i \in [t-K, t+K]} E_i.
\end{equation}

\subsection{Closed-Loop Atomic Domain Generation} \label{sec:domain_generation}

Given an ordered keyframe sequence $\{k_1, \ldots, k_N\}$ and the associated task instruction \instruct, we construct a symbolic PDDL domain through a multi-stage LLM-guided pipeline (Figure~\ref{fig:overview}b).  
For each transition $(k_i, k_{i+1})$, a vision-language model (VLM) infers the operator name, identifies preconditions and effects, and expands the predicate set if necessary, yielding an initial grounded domain $\domain_0$. To improve consistency and generality, we pass $\domain_0$ and \instruct through a large language model (LLM) for holistic revision. The LLM enforces syntactic correctness, predicate reuse, and naming consistency, producing a revised domain $\domain_r$.
We then apply two nested verification steps to refine the domain:

\textbf{Solvability Check.}  
To assess domain correctness, we prompt the LLM with $(\prset_r, \instruct)$ to generate $K$ test problems $\{\problem_1, \ldots, \problem_K\}$ of increasing difficulty. Only the predicate set $\prset_r$ is used for problem generation, preventing the LLM from compensating for incorrect operators in $\domain_r$. Each pair $(\domain_r, \problem_k)$ is evaluated by a PDDL planner. The solvability score is defined as
\begin{equation}
S(\domain_r) = \frac{1}{K}\sum_{k=1}^{K} \mathbb{I}\left[ \text{PDDLSolver}(\domain_r, \problem_k) \text{ solves } \problem_k \right].
\label{eq:solvability-score}
\end{equation}
If $S(\domain_r) < \theta$ (default $K=5$, $\theta=0.6$), the full, verbose feedback from the PDDL planner, including search process logs and any validation errors, is passed back to prompt the LLM for domain refinement, resulting in an updated atomic domain $\domain_s$.

\textbf{Solution Verification.}  
We then verify the solution to the solvable and most challenging test problem, $\actseq_K = \text{PDDLSolver}(\domain_s, \problem_K)$, using another LLM to check whether the plan satisfies physical constraints and commonsense expectations. The LLM reads the action sequence and identifies steps that violate physical or operational commonsense. For example, it flags errors such as trying to pick an occluded object, stacking an object on itself, or applying an unsupported action to objects. If any violations are found, LLM feedback is used to prompt further domain refinement. 

The two nested checks are repeated until both pass or a maximum of $L=5$ iterations is reached. If convergence fails, the learned atomic domain is discarded. The entire closed-loop process can be restarted to regenerate the domain.

\subsection{The \Alldomains}

% \paragraph{Scaling to Large Demonstration Sets.}  
Using the domain learning pipeline described above, we process a total of \textbf{12,393} demonstrations from DROID, each yielding a corresponding \singledomain. While each \singledomain $\domain_i$ captures task-specific knowledge grounded in a single demonstration, the complete set forms a comprehensive \alldomains that spans the full \taskspace present in the dataset.
% \paragraph{A Symbolic Knowledge Graph View.}  
Each \singledomain can be interpreted as a minimal symbolic \knowledgegraph $\domain = (\mathcal{V}, \mathcal{E})$, where the vertex set $\mathcal{V} = \prset \cup \oprset$ includes predicates $\pre \in \prset$ and operators $\opr \in \oprset$, and the edge set $\mathcal{E} = \mathcal{E}_{\text{pre}} \cup \mathcal{E}_{\text{eff}}$ encodes preconditions ($\pre \xrightarrow{\text{pre}} \opr$) and effects ($\opr \xrightarrow{\text{eff}} \pre$). 
Aggregating all \singledomains (by taking the union of predicate and operator sets and merging directly-overlapping nodes), the \alldomains forms a large-scale symbolic \knowledgegraph for real-world robotic task planning (Figure~\ref{fig:Knowledge Graph}), comprising \textbf{3,137} operator nodes grouped into \textbf{170} semantic categories, \textbf{2,875} predicate nodes, and \textbf{16,481} causal edges.  
This unified representation connects otherwise isolated behaviors such as \texttt{pick\_from\_table}, \texttt{pour\_into\_bowl}, \texttt{stir\_bowl}, \texttt{remove\_lid}, \texttt{place\_in\_pot}, and \texttt{wipe\_table}, enabling the planner to solve long-horizon tasks like ``\texttt{heat the milk using the pot and clean the table}'' through compositional generalization across \singledomains.

% \paragraph{Challenges in Using the Raw Graph.}  
Despite its broad coverage, the \alldomains is not directly suitable for online planning. Its large scale poses challenges for LLMs and VLMs, making task grounding less reliable and increasing computational overhead. 
Second, semantic inconsistencies across \singledomains---such as varying predicate names for equivalent concepts---break symbolic continuity and reduce planning effectiveness.  
Therefore, to ensure the quality of domains and enable effective planning, for a specific \taskgroup, we first retrieve a relevant subset of \singledomain{}s based on the relevance of language instructions, and then fuse them into a compact, high-quality \fuseddomain. 

\section{Domain Fusion}

With a retrieved set of task-relevant \singledomains, $\{\domain_i\}_{i=1:M}$, this section constructs a \fuseddomain, $\domain = \bigcup_i \domain_i$, an integrated graph with improved symbolic and causal connectivity, so that it better supports generalization across task variations. This is achieved by merging functionally-overlapping nodes, via hierarchical fusion along a binary tree (Figure \ref{fig:overview}c).
% Note that a naïve union is not sufficient here due to semantic inconsistency, and would lead to very low planning success rate in out-of-distribution tasks. Instead, we construct a \fuseddomain via hierarchical fusion along a binary tree (Figure \ref{fig:overview}c). 

% To ensure consistency and compositionality, we adopt a hierarchical fusion strategy that recursively merges domain pairs along a binary tree structure. This bottom-up process incrementally builds a coherent and compact \fuseddomain from initially disjoint \singledomains. The algorithmic details of this fusion method are described in the next section.

% We treat each PDDL domain $\domain = (\mathcal{V}, \mathcal{E})$ as a symbolic knowledge graph, where the vertex set $\mathcal{V} = \prset \cup \oprset$ contains predicates $\pre \in \prset$ and operators $\opr \in \oprset$, and the edge set $\mathcal{E} = \mathcal{E}_{\text{pre}} \cup \mathcal{E}_{\text{eff}}$ encodes causal relationships: preconditions ($\pre \xrightarrow{\text{pre}} \opr$) and effects ($\opr \xrightarrow{\text{eff}} \pre$). 
% This structure captures dependencies among actions in task execution. While each individual domain $\domain_i$ encodes task-specific knowledge, they cannot generalize to new tasks due to inconsistent naming and disconnected symbol structures. 
% Our key insight is that fusing a diverse collection $\{\domain_1, \ldots, \domain_n\}$ into a unified symbolic graph enables compositional generalization across domains.

\subsection{Atomic Domain Retrieval}
The set of task-relevant atomic domains can be retrieved either manually or automatically. For manual retrieval, atomic domains can be selected based on whether the associated language instructions show relevance to the target task class. For automatic retrieval,  we prompt an LLM to infer the set of relevant actions based on the language description of the target task class. We then use sentence embedding similarity to find the top-K matching operators in the unified domain. Atomic domains containing these relevant operators are thus retrieved for fusion.

\subsection{Binary Tree Fusion}
We construct the \fuseddomain{} by recursively merging a set of \singledomains{} along a binary tree. At each level $l$, a parent node $\domain_k^l$ is formed by fusing its two arbitrarily-paired child domains:
$
\domain_k^l = f(\domain_{2k-1}^{l+1}, \domain_{2k}^{l+1})$,
where $f(\cdot, \cdot)$ performs structured alignment of predicates and operators. This process proceeds bottom-up until a single, unified \fuseddomain{} resides at the root. 
Each node fusion is performed in two stages:

\textbf{Predicate Merging.}  
Let $\prset_1$ and $\prset_2$ be the predicate sets from two child domains. We compute predicate similarity using $\phi(\pre_i, \pre_j) = \cos(\mathrm{E}(\pre_i), \mathrm{E}(\pre_j))$, where $\mathrm{E}(\cdot)$ denotes a text embedding from a pretrained language model~\cite{mpnet}. Predicate pairs with $\phi < \tau_{p}$ (with $\tau_{p} = 0.3$) are discarded. The remaining candidates are ranked by similarity and sequentially verified by an LLM for semantic equivalence. Equivalent predicates are merged.

\textbf{Operator Merging.}  
Following predicate alignment, we update all operators referencing merged predicates using LLM assistance. Operator similarity is computed via name embeddings: $\phi_{\text{name}}(\opr_1, \opr_2) = \cos(\mathrm{E}(\texttt{name}(\opr_1)), \mathrm{E}(\texttt{name}(\opr_2)))$. Pairs with $\phi_{\text{name}} < \tau_{o}$ (with $\tau_{o} = 0.3$) are filtered out. The remaining operator pairs are ranked and passed to the LLM along with their $(\texttt{name}, \texttt{pre}, \texttt{eff})$ tuples. Functionally equivalent ones are merged, inheriting the union of their preconditions and effects.

\section{Task Planning with \algname}

Given a new task specified by a language instruction \instruct and a scene image \image, we apply the \fuseddomain \domain for symbolic planning. The pipeline consists of two stages: (1) constructing a task-specific PDDL problem \problem, and (2) solving the pair $(\domain, \problem)$ using a classical planner such as Fast Downward~\cite{fastdownward}. A central challenge is filtering task-relevant elements from \domain to reduce symbolic noise and improve solver efficiency. 

\textbf{Predicate Grouping.}  
To help the LLM interpret the large predicate set \prset in \domain, we pre-organize predicates into four semantic groups: object category descriptors, state or attribute indicators, spatial relations, and affordance-related predicates. This structured input improves the reliability of downstream problem construction.

\textbf{Predicate and Operator Filtering.}  
We first prompt a vision-language model with $(\domain, \image, \instruct)$ to generate an initial problem: $\problem_0 = \text{LLM}(\domain, \image, \instruct)$.
From $\problem_0$, we extract the predicate set $\prset_0$ used in the initial and goal conditions, treating these as task-relevant predicates. Next, we extract from \domain the operators whose preconditions or effects involve 
\begin{equation}
    \prset_0: \oprset_{\text{pre}} = \{ \opr \in \oprset : \exists \pre \in \prset_0, \, \pre \xrightarrow{\text{pre}} \opr \}, \oprset_{\text{eff}} = \{ \opr \in \oprset : \exists \pre \in \prset_0, \, \opr \xrightarrow{\text{eff}} \pre \},
\end{equation}
and define the reduced operator set 
\begin{equation}
    \oprset' = \oprset_{\text{pre}} \cup \oprset_{\text{eff}}.
\end{equation}
Using $\oprset'$, a compact domain $\domain_{\text{new}} = (\prset_0, \oprset')$ is constructed, and a refined problem is generated:
$\problem_{\text{new}} = \text{LLM}(\domain_{\text{new}}, \image, \instruct)$.

\textbf{PDDL Planning.}  
We then solve $(\domain, \problem_{\text{new}})$ using a symbolic planner,
$\actseq = \text{PDDLSolver}(\domain, \problem_{\text{new}})$,
yielding an action sequence \actseq that satisfies \goal under the symbolic constraints and minimizes cost. This filtering process improves both planning accuracy and computational efficiency by reducing irrelevant symbolic clutter.

\section{Experiments}
We evaluate UniDomain on diverse real-world tasks. Results demonstrate that UniDomain achieves substantial improvements over the strongest baseline methods, obtaining up to 58\% higher task success rate and 160\% higher plan optimality. Specifically, UniDomain maintains consistently high success and optimality across diverse and previously unseen tasks, significantly reducing planning overhead compared to both LLM-only planners (e.g., Code-as-Policies \cite{cap} and ReAct \cite{react}) and hybrid LLM-PDDL methods (e.g., ISR-LLM \cite{isr-llm} and BoN-iVML \cite{BoN-iVML}). Further ablation studies confirm that these performance gains result from learning a comprehensive domain via closed-loop verification and structured fusion and effectively filtering irrelevant predicates and operators during planning. 

\subsection{Experimental Setup}

\textbf{Tasks.}
The evaluation tasks span 4 unseen task domains: \textit{BlockWorld}, \textit{Desktop}, \textit{Kitchen}, and \textit{Combination}. \textit{BlockWorld} involves block sorting and stacking with ordering constraints; \textit{Desktop} includes drawer use, wiping, folding, and document organization; \textit{Kitchen} covers object transfers and food-tool manipulation; \textit{Combination} mixes all domains to test cross-context generalization.  There are 100 tasks in total. 40 atomic domains learned from DROID demonstrations are retrieved to construct a \fuseddomain for all evaluation tasks, which includes {78} predicates and {61} operators. See details of the evaluation tasks and the \fuseddomain in Appendix \ref{apx:tasks} and Appendix \ref{apx:verb}, respectively.

\textbf{Evaluation Metrics.}
We report success rate \textit{(SR)}, success-weighted relative path length (\textit{SPL}), and optimality rate (\textit{OR}), defined as the fraction of plans whose cost $c_i$ falls within a threshold $K$ of the optimal $c_i^*$. Specifically,
\begin{equation}
\text{SPL} = \frac{1}{N} \sum_{i=1}^{N} \mathbb{I}_{\text{succ}} \cdot \frac{c_i^*}{c_i},
\end{equation}

\begin{equation}
\text{OR}(K) = \frac{1}{N} \sum_{i=1}^{N} \mathbb{I}\left[0 < c_i \leq c_i^* + K \right].
\end{equation}

We additionally report the thinking time (LLM wall-clock runtime) and number of LLM calls per task to assess efficiency and overhead.

\textbf{Baselines.}
We compare \algname against two categories of methods. The first uses LLMs or VLMs \textit{as planners}:
\textit{Code-as-Policies} \cite{cap} directly generates executable Python-style plans from language instructions;
\textit{ReAct} \cite{react} improves robustness through closed-loop reasoning with feedback;
\textit{VLM-CoT} applies chain-of-thought prompting \cite{cos} in a zero-shot vision-language setting.
The second category integrates LLMs with PDDL planning:
\textit{ISR-LLM} \cite{isr-llm} translates instructions into PDDL specifications for building LLM planning and iteratively refines plans with validator feedback;
\textit{VLM-PDDL} grounds scene and language into symbolic specifications and plans with classical solvers;
\textit{BoN-iVML} \cite{BoN-iVML} generates an initial PDDL domain via Best-of-N sampling, refines it with verbalized feedback, and then constructs the problem file for planning.

\textbf{Evaluation Protocol.}
We evaluate UniDomain as a high-level task planner, not as an integrated robot system. To focus evaluation on the performance of high-level symbolic planning, we followed a standard practice in the task planning literature \cite{robovqa, embodied-reasoner} and assumed a perfect low-level control policy (human teleoperation in our experiments for both UniDomain and baselines), so that the measured performance does not get confounded with potential imperfections in the low-level controller. We used a semi-automatic evaluation approach, wherein an LLM reads the task and the plan to provide an initial assessment, followed by final verification by human experts. Example results in Appendix \ref{apx:eval} show that these judgments reflect well-defined objectives and commonsense constraints.

Despite the stand-alone evaluation, \algname is ready for seamless integration into a complete robotic system. Its high-level plan can be straightforwardly translated into natural language commands and input to any low-level language-conditioned skill policy, like modern Vision-Language-Action (VLA) models \cite{rdt, pi0}, or modular approaches combining perception, motion planning, and affordance learning. See real-world demonstrations of such an integrated system built upon \algname on our project website: \url{https://roboticsjtu.github.io/UniDomain/}.

\subsection{Comparison Results} \label{sec:comparisons}
\begin{figure}[t!]
\centering
    % 图例整体一行，左图例正常，右图例上移
    \makebox[\textwidth][c]{%
        \begin{minipage}[t]{0.6\linewidth}
            \hspace*{0.2cm}  % 微调左图例位置
            \includegraphics[width=\linewidth]{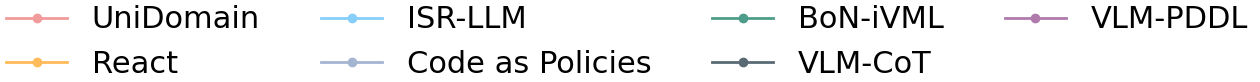}
        \end{minipage}%
        \hfill
        \raisebox{0.28cm}{  % 单独上移右图例
            \begin{minipage}[t]{0.35\linewidth}
                \hspace*{-0.2cm}  % 微调右图例位置
                \includegraphics[width=\linewidth]{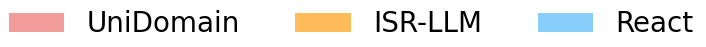}
            \end{minipage}
        }%
    }

    % 三个子图
    \begin{tabular}{@{\hskip 2pt}c@{\hskip 2pt}c@{\hskip 2pt}c@{\hskip 2pt}}
        \includegraphics[width=0.6\linewidth]{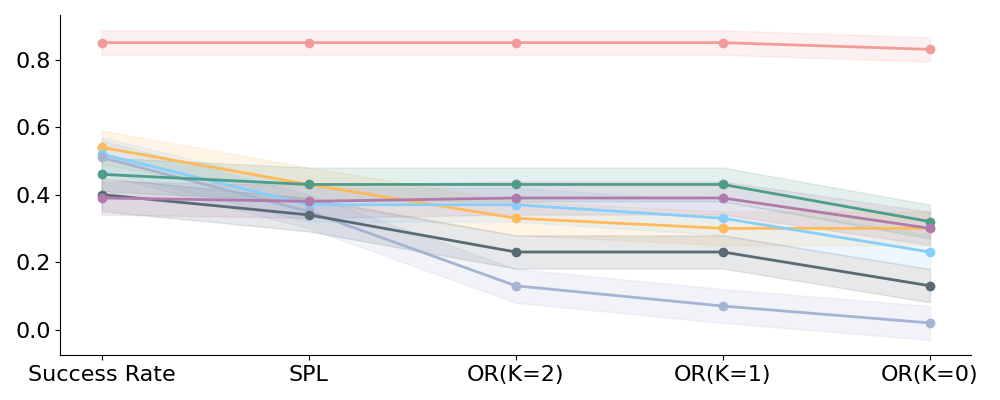}  &
        \raisebox{0.2cm}
        {\includegraphics[width=0.15\linewidth]{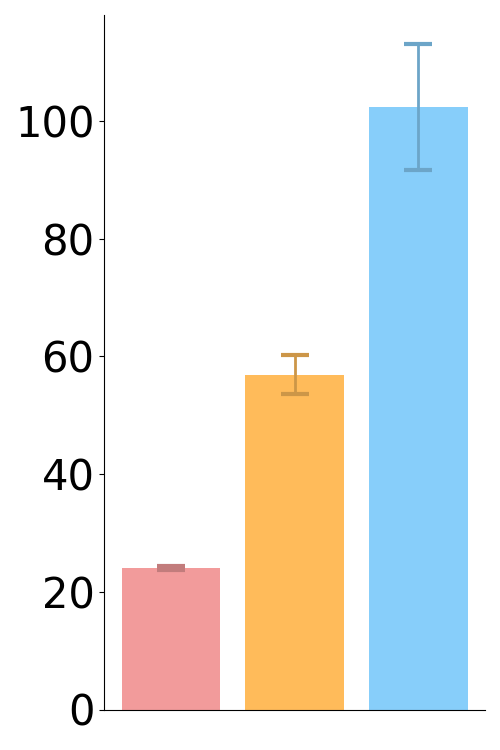}} & 
        \raisebox{0.2cm}
        {\includegraphics[width=0.15\linewidth]
        {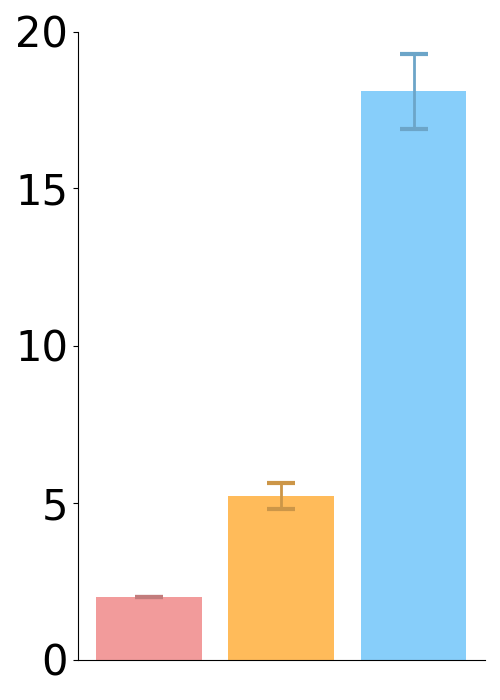}} \\
        \small{(a)} &  \small{(b)}  &   \small{(c)} \vspace{-0.5em}
    \end{tabular}
    \vspace*{0.4cm}
    \caption{Comparison results of \algname and state-of-the-art methods on unseen evaluation tasks: 
    (a) success rates $\uparrow$, success-weighted relative path lengths $\uparrow$, and optimality rates with thresholds ($K=2,1,0$) $\uparrow$; 
    (b) Thinking time (s) $\downarrow$ of the top-performing methods; 
    (c) number of LLM calls $\downarrow$ of the top-performing methods. 
    Average values are shown with standard errors.}
    \label{fig:comparison}
    \vspace*{-0.3cm}
\end{figure}

Figure~\ref{fig:comparison}a reports performance across three metrics: \textit{Success Rate}, \textit{SPL}, and \textit{Optimality Rate} at increasing strictness levels ($K{=}2,1,0$), transitioning focus from task feasibility to plan optimality. All methods were evaluated using GPT-4.1 via API under a fixed temperature of 0.0.

Among LLM-only planners, \textit{Code-as-Policies} achieves moderate success (51\%) but degrades rapidly under stricter thresholds, highlighting its limited global reasoning capacity. \textit{VLM-CoT} produces slightly more optimal plans due to stronger visual grounding, but struggles with task completion due to the absence of symbolic structure. \textit{ReAct} yields the highest success among LLM-only methods by leveraging action feedback, yet its lack of explicit state tracking results in redundant or invalid steps—especially on long-horizon tasks.

For PDDL-integrated planners, \textit{VLM-PDDL} performs comparably to \textit{VLM-CoT}, but is hindered by fragile domain and problem generation—minor grounding or typing errors often lead to unsolvable plans. \textit{ISR-LLM} achieves the highest task success rate among all baselines through iterative validator feedback, but its reliance on LLM-based planning results in sharp optimality drops (matching \textit{VLM-CoT} at $K{=}0$). \textit{BoN-iVML} improves over \textit{VLM-PDDL} with verbalized refinement but still fails to construct reliable high-quality domains on the fly.

In contrast, \algname achieves strong and consistent performance across all metrics, attaining 85\% success rate and 83\% optimality at $K{=}0$. It also incurs the lowest LLM thinking time and fewest LLM calls among top-performing methods (Figures~\ref{fig:comparison}b and~\ref{fig:comparison}c). 
The \fuseddomain effectively supports compositional generalization. In $83\%$ of tasks, our planner successfully produced not only feasible but also optimal plans through the composition of learned operators in \alldomains. These results show that pre-training the \alldomains, paired with post-training using domain fusion and test-time problem generation, enables symbolic planning to outperform both end-to-end LLM agents and the best existing hybrid planners in robustness, efficiency, and plan quality.

\subsection{Ablation Studies}
We conduct ablation studies to understand the contributions of core components in \algname.
Results show that removing the closed-loop verification significantly reduces \singledomain quality, causing failures in solvability and task logic. Hierarchical fusion is critical, as a naive union of \singledomains or direct LLM-based merging yields unusable domains due to semantic and structural inconsistencies. Additionally, predicate grouping and task-relevant filtering substantially boost planning performance, particularly in tasks requiring complex reasoning and compositional generalization.

% a figure with 3 subfigures (each for one type of ablations)

\begin{figure}[t!]
    \vspace*{0.2cm}
    \centering
    \begin{tabular}{@{\hskip 2pt}c@{\hskip 20pt}c@{\hskip 2pt}}
        \begin{overpic}[width=0.4\linewidth]{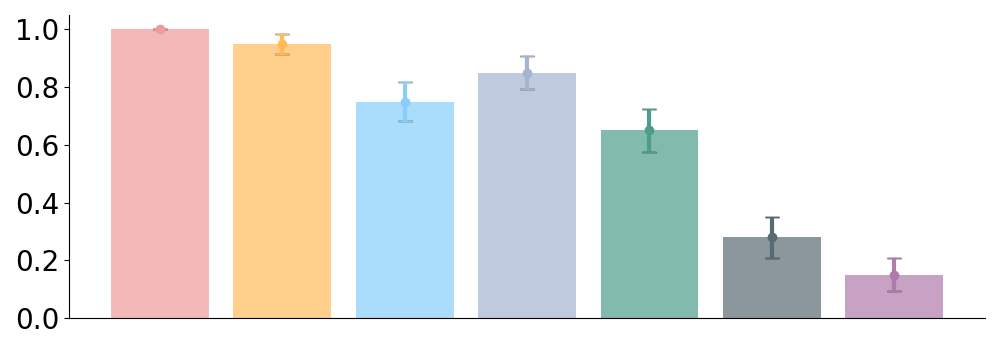}
            \put(-5,38){\includegraphics[width=0.45\linewidth]{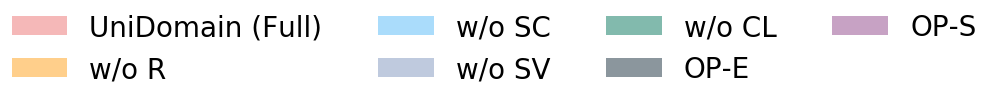}}
        \end{overpic} &
        % 使用 raisebox 调整图(b)的整体垂直位置
        \raisebox{-2mm}{
        \begin{overpic}[width=0.5\linewidth]{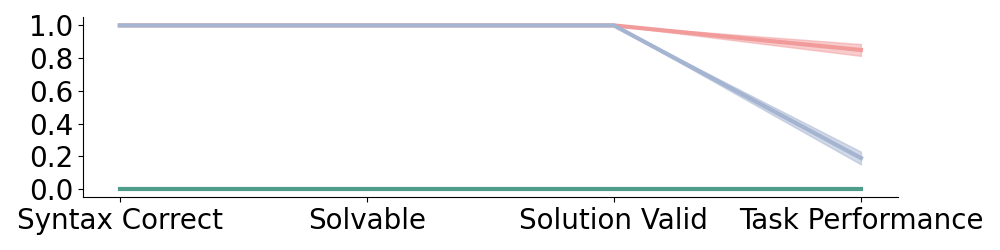}
            \put(0,37.5){\includegraphics[width=0.5\linewidth]{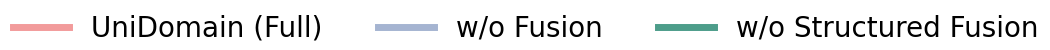}}
        \end{overpic}} \vspace{0.3em} \\
        (a) &   (b) \vspace{-0.5em}
    \end{tabular}
    \vspace*{0.2cm}
    \caption{Results for ablation studies on domain generation: (a) ablation on the \singledomain learning method; (b) ablation on the domain fusion method. All values are success rates $\uparrow$ with standard errors.}
    \label{fig:ablation_learning}
\end{figure}

\paragraph{Ablations on Domain Learning.}In Figure~\ref{fig:ablation_learning}a, we perform ablation studies on our atomic domain learning pipeline using 40 DROID demonstrations with paired instructions. A domain is considered successful if it passes both the solvability check, ensuring that test problems can be solved by a PDDL planner, and the plan verification step, which checks whether the resulting plans conform to real-world physical constraints and commonsense logic. We report the success rate as the primary metric.

To assess the contribution of each module, we compare against several ablated variants. Removing the LLM-based revision step (\textit{w/o R}) results in domains with more syntax or logical inconsistencies, increasing the average number of required refinement iterations (from 0.49 to 1.36). Disabling the solvability check (\textit{w/o SC}) leads to domains that are syntactically valid but often fail due to disconnected operators and incomplete predicates. Removing the solution verification stage (\textit{w/o SV}) produces domains that are solvable but fail to capture essential task logic, resulting in mis-aligned plans. Eliminating all feedback mechanisms (\textit{w/o CL}) reduces the process to single-pass LLM generation, which significantly degrades domain quality.

We also evaluate one-shot variants that use no refinement or validation, generating the domain from a single LLM query. Using our energy-based keyframe extraction (\textit{OP-E}) achieves 28\% single-pass success, while the similarity-based approach~\cite{similarity} (\textit{OP-S}) drops to 15\%. In addition to higher accuracy, our energy-based method is also substantially more efficient, reducing the processing time per demonstration from 47.8 seconds to just 0.6 seconds on average. \footnote{We performed similarity-based keyframe extraction using SigLIP-2 \cite{siglip2} on an NVIDIA A800 GPU (80GB VRAM), running in parallel across batches. The energy-based method was executed in single-threaded mode on an i7-14700HX CPU (32GB RAM).} 
\vspace*{-0.2cm}
\paragraph{Ablations on Domain Fusion.}
% \begin{table}[!t]
% \centering
% \caption{Ablation results on domain components. Metrics include syntax validity, solvability, commonsense check pass rate, and task success rate.}
% \begin{tabular}{lcccc}
% \toprule
% Method & Syntax Validity & Solvability & Plan Validity & Success Rate \\
% \midrule
% UniDomain & 1.00 & 1.00 & 1.00 & 0.85 \\
% w/o Domain Fusion & 1.00 & 1.00 & 1.00 & 0.19 \\
% w/o Binary Tree Fusion & 0.00 & 0.00 & 0.00 & 0.00 \\
% \bottomrule
% \end{tabular}
% \label{tab:ablation}
% \end{table}
In Figure~\ref{fig:ablation_learning}b, we perform ablation studies on our hierarchical domain fusion method using test problems generated from evaluation tasks (sampled from those in Section \ref{sec:comparisons}), assessing whether the fused domain supports correct and generalizable planning. We report four metrics: syntax validity (as measured by a PDDL syntax verifier), solvability (the rate of passing solvability tests), solution validity (the rate of passing plan verification), and task performance measured as the success rate of online planning.

Our full method, UniDomain, achieves perfect scores on syntax, solvability, and verification, and attains 85\% task success. In contrast, using \singledomains without fusion (\textit{w/o Domain Fusion})---where the planner selects the closest \singledomain for each task, akin to retrieval-based methods---yields only 19\% success, despite perfect syntax and solvability. 
This confirms the power of compositional generalization.
% This demonstrates that local domains lack the coverage and compositionality needed for generalization. 
Furthermore, replacing our structured fusion with a direct LLM-based merging strategy (\textit{w/o Structured Fusion}) fails entirely: the merged domains contain structural errors, violate syntax rules, and are unusable by downstream planners.

% These results highlight that hierarchical fusion of predicates and operators is essential for building a coherent, planner-compatible domain that generalizes across tasks. Naïve LLM-based merging fails to preserve logical structure, and domain selection alone lacks combinatorial expressivity.

% \subsubsection{Ablations on Domain Fusion}

% To examine the role of the structured binary-tree-based domain fusion, we conduct an ablation (\textbf{w/o Domain Fusion}) in which the LLM directly selects a single domain for planning based solely on the instruction and scene image, bypassing the explicit fusion of individual domains. This ablation leads to a pronounced decline in performance.
% The degradation arises because removing fusion eliminates the systematic integration of complementary operators and predicates spread across multiple domains, which is essential for addressing complex, multi-skill tasks. Notably, limited success under this condition is observed only in the BlockWorld environment, where one of the original single domains happens to cover the entire task space. In contrast, the remaining environments involve inherently composite tasks, whose required operators are distributed across different domains; no single domain can independently support complete task execution. As a result, without domain fusion, the system fails entirely in these cases. 
% This highlights a fundamental limitation: explicit domain fusion is indispensable for constructing the cross-domain operator and predicate combinations required for solving diverse and generalizable planning problems.

\paragraph{Ablations on the Planning Method.}

% \begin{wrapfigure}{r}{0.5\textwidth}
\begin{figure}[t!]
    \centering
    \includegraphics[width=0.55\linewidth]{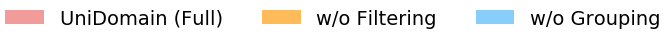} 
    \includegraphics[width=0.55\linewidth]{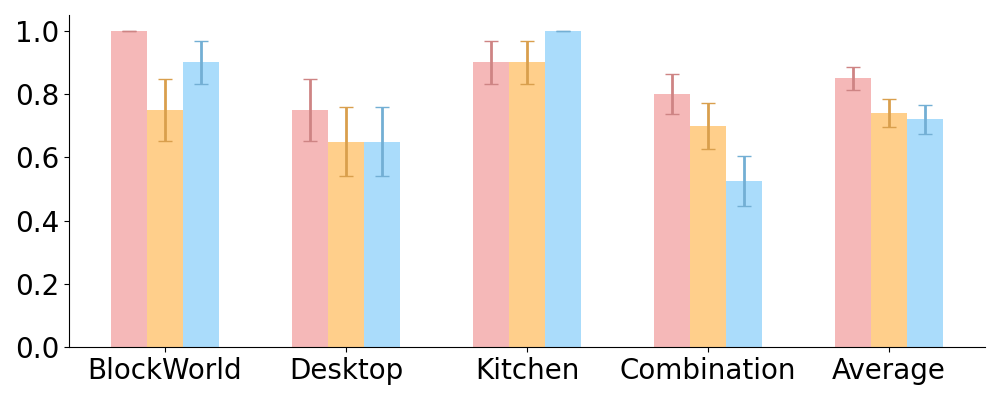}
    \caption{Results for ablation study of the \algname planner. Each bar shows average task success rates $\uparrow$ with standard errors.}
    \label{fig:ablation_planning}
\end{figure}
% \end{wrapfigure}

In Figure ~\ref{fig:ablation_planning}, we assess the impact of predicate organization and domain filtering on planning performance by measuring task success rates across all evaluation tasks. In \textit{w/o Grouping}, we remove the structural organization of predicates into semantic categories, providing the full flat list to the LLM during problem generation. In \textit{w/o Filtering}, we disable pruning based on task relevance and directly use the full \fuseddomain to perform a single-pass problem generation.

Removing predicate grouping degrades performance significantly, particularly in the \textit{Combination} domain, where complex task composition requires the LLM to interpret a rich and diverse predicate space. Without grouping, the flat structure overwhelms the LLM's capacity to localize task-relevant semantics. Disabling predicate and operator filtering also leads to sharp performance drops, especially in the \textit{BlockWorld} domain. These tasks rely on long-horizon action dependencies, and a compact, task-focused domain allows more coherent grounding and reasoning by reducing irrelevant information. 

\section{Conclusion and Limitations}

We present \algname, a framework that addresses the challenge of task planning under complex, implicit constraints from language and vision. \algname learns a reusable PDDL domain from large-scale visual demonstrations and applies it to zero-shot symbolic planning. By combining closed-loop domain learning, hierarchical fusion, and task-relevant filtering, \algname enables efficient and generalizable planning across diverse tasks. Experiments on 100 real-world \tasks demonstrate that \algname substantially outperforms prior LLM-only and hybrid LLM-PDDL baselines, achieving higher success rates and plan optimality.

Despite its strong performance, \algname has a few limitations.
First, automatically-retrieved atomic domains can be redundant, thus construction of a \fuseddomain{} can be time-consuming. Future work will focus on improving the accuracy and efficiency of the domain retrieval and fusion methods. Second, \algname operates under the PDDL 1.0 formalism, which lacks support for temporal constraints, numeric fluents, and cost-sensitive planning. Extending the framework to richer representations such as PDDL 2.1~\cite{pddl2.1} is an important direction.
Finally, our experiments assume full observability, ignoring real-world challenges like occlusion and perceptual noise. Incorporating probabilistic planning frameworks such as PPDDL~\cite{ppddl} or RDDL~\cite{rddl} is a promising path toward handling uncertainty.

\section*{Acknowledgements}
This work was supported in part by the National Key R\&D Program of China (Grant No. 2024YFB4707600) and National Natural Science Foundation of China under grant No. 62303304.

We used generative AI to improve self-written texts to enhance readability. None of the presented
methods and results (figures, equations, numbers, etc.) are generated by AI.
% None of the presented methods and results (figures, equations, numbers, etc.) are generated by AI.

% \section*{References}
\newpage
\bibliographystyle{unsrt} % 可选样式: plain, abbrv, unsrt, alpha 等
\bibliography{references}

\newpage
\appendix
\begin{center}
    {\LARGE\bfseries Appendix}
\end{center}
\section{The Task Included in Overview}\label{apx:overview}
\begin{figure}[h!]
    \centering
    \includegraphics[width=0.9\linewidth]{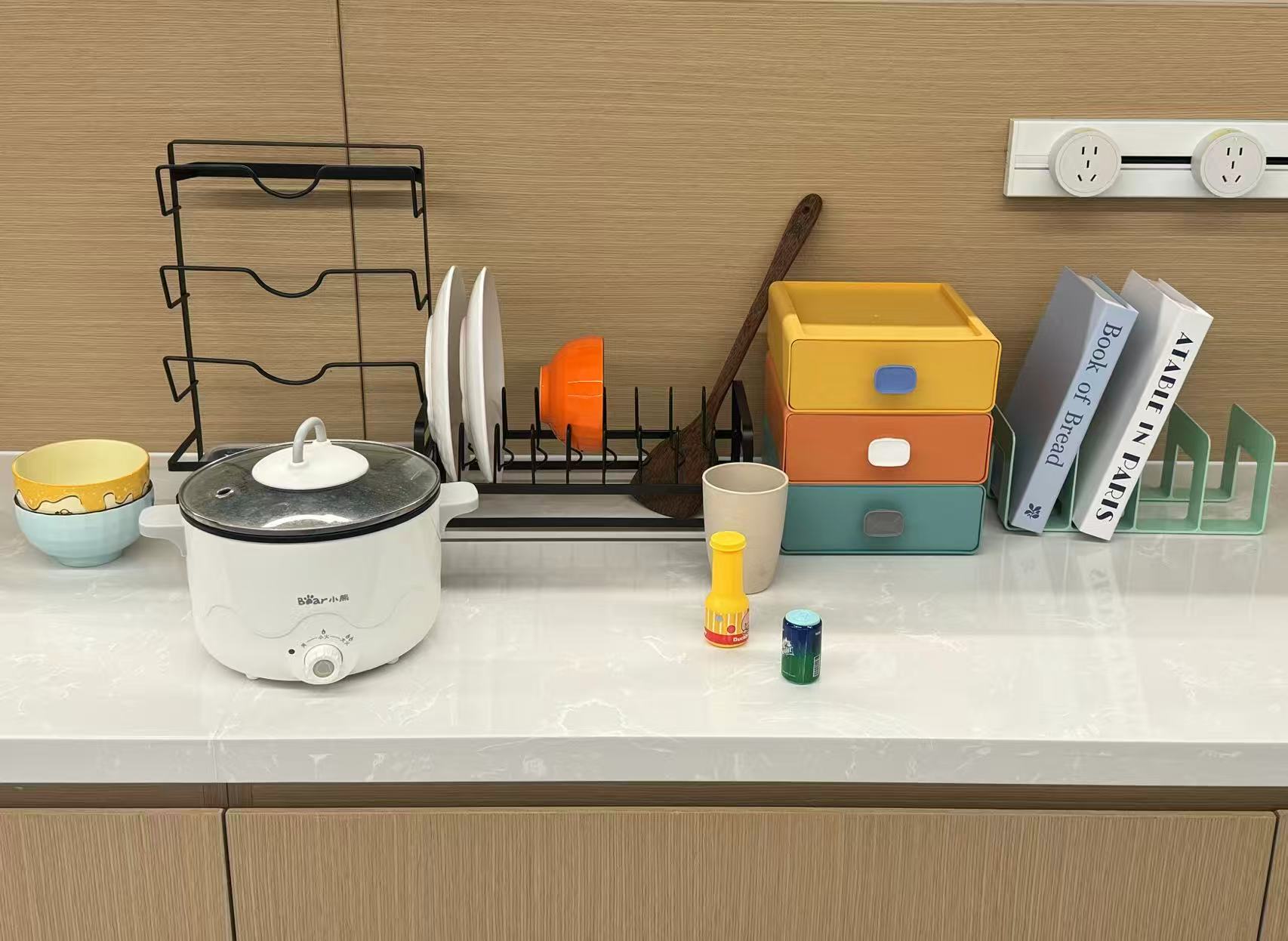}
    \caption{The initial scene of the task in the overview (Figure \ref{fig:overview}).}
    \label{fig:init_scene}
\end{figure}

The task scene used in the overview is shown in Figure \ref{fig:init_scene}. The language instruction is \textit{``Move the corn from the pot into the orange bowl, wipe the table with the towel in the drawer and put it back to the closed drawer.''} The meta-domain used is fused from 40 atomic domains learned from DROID demonstrations, and includes key operators for accomplishing such tasks, 
% like \texttt{remove\_lid}, \texttt{pick\_from\_rack}, \texttt{pick\_from\_pot}, \texttt{place\_on\_table}, \texttt{put\_in\_bowl}, and \texttt{wipe\_table}
for example:
\begin{lstlisting}[basicstyle=\ttfamily\small]
(:action pick_from_rack
    :parameters (?r ?o ?rk)
    :precondition (and (on_rack ?o ?rk) (hand_free ?r) (rack ?rk))
    :effect(and(not(on_rack ?o ?rk))(not (hand_free?r))(holding ?r?o))
)
(:action place_on_table 
    :parameters (?r ?o ?t) 
    :precondition (and (holding ?r ?o) (table ?t)) 
    :effect (and(on_table ?o ?t)(hand_free ?r)(clear ?o)(not (holding ?r ?o))))
(:action pick_from_pot
    :parameters (?r ?o ?p)
    :precondition (and (in_pot ?o ?p) (is_open ?p) (hand_free ?r))
    :effect(and(holding ?r ?o)(not(in_pot ?o ?p))(not(hand_free ?r)))
)
(:action open_drawer 
     :parameters (?r ?x) 
     :precondition (and (hand_free ?r) (drawer ?x) (not (is_open ?x))) 
     :effect (and (is_open ?x))
)
\end{lstlisting}

The online planner generates the following PDDL problem from the language and image input (showing only core predicates):
\begin{lstlisting}[basicstyle=\ttfamily\small]
(:init 
    (on_rack orange_bowl rack)
    (in_pot corn pot)
    (can_wipe_table towel)
    (in_drawer towel yellow_drawer)
    (hand_free robot)
    (on lid pot)
)
(:goal 
    (and 
        (in_bowl corn orange_bowl) 
        (wiped table) 
        (in_drawer towel yellow_drawer) 
        (not (is_open yellow_drawer))
    )
)
\end{lstlisting}

\begin{figure}[h!]
    \centering
    \includegraphics[width=0.9\linewidth]{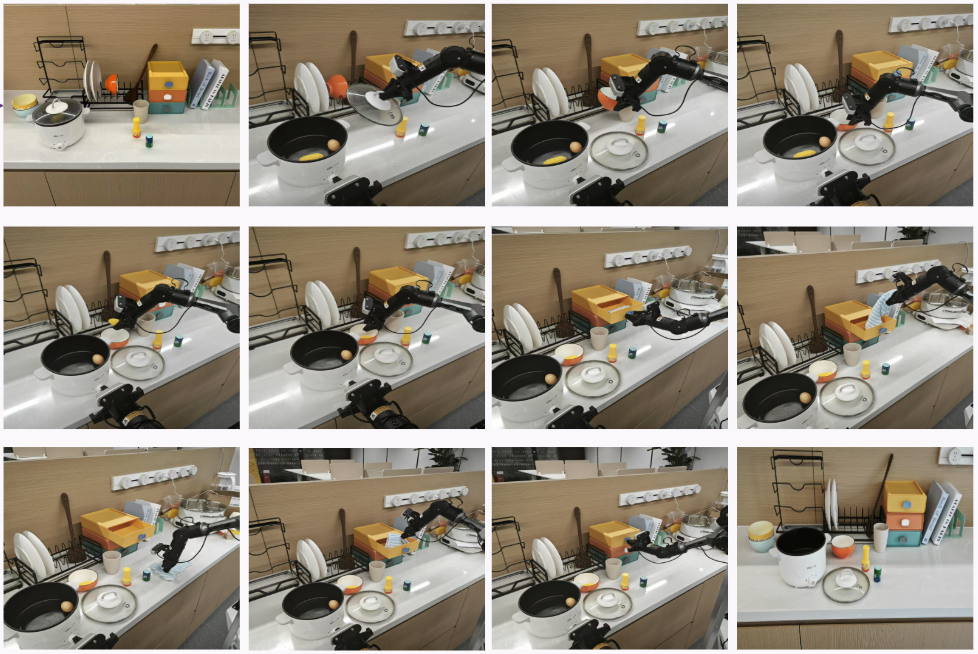}
    \caption{Execution of the high-level plan generated by UniDomain.}
    \label{fig:overview_task}
\end{figure}

The PDDL solver produces the following optimal plan:

\begin{lstlisting}[basicstyle=\ttfamily\small]
(remove_lid robot lid pot)
(pick_from_rack robot orange_bowl rack)
(place_on_table robot orange_bowl table)
(pick_from_pot robot corn pot)
(put_in_bowl robot corn orange_bowl table)
(open_drawer robot yellow_drawer)
(pick_from_drawer robot towel yellow_drawer)
(wipe_table robot towel table)
(place_in_drawer robot towel yellow_drawer)
(close_drawer robot yellow_drawer)
\end{lstlisting}
Visual results of the plan execution are shown in Figure \ref{fig:overview_task}.

\newpage
\section{Evaluation Tasks}\label{apx:tasks}
We provide representative examples of task images and corresponding language instructions for four real-world task domains—\textit{BlockWorld}, \textit{Desktop}, \textit{Kitchen}, and \textit{Combination}—each illustrated with two representative task instances used in our experimental evaluation.

\subsection{Representative Tasks in \textit{BlockWorld}}\label{apx:bw}

\begin{center}
%----------- Row 1 -----------
\begin{minipage}[t]{0.45\textwidth}
  \includegraphics[width=\linewidth]{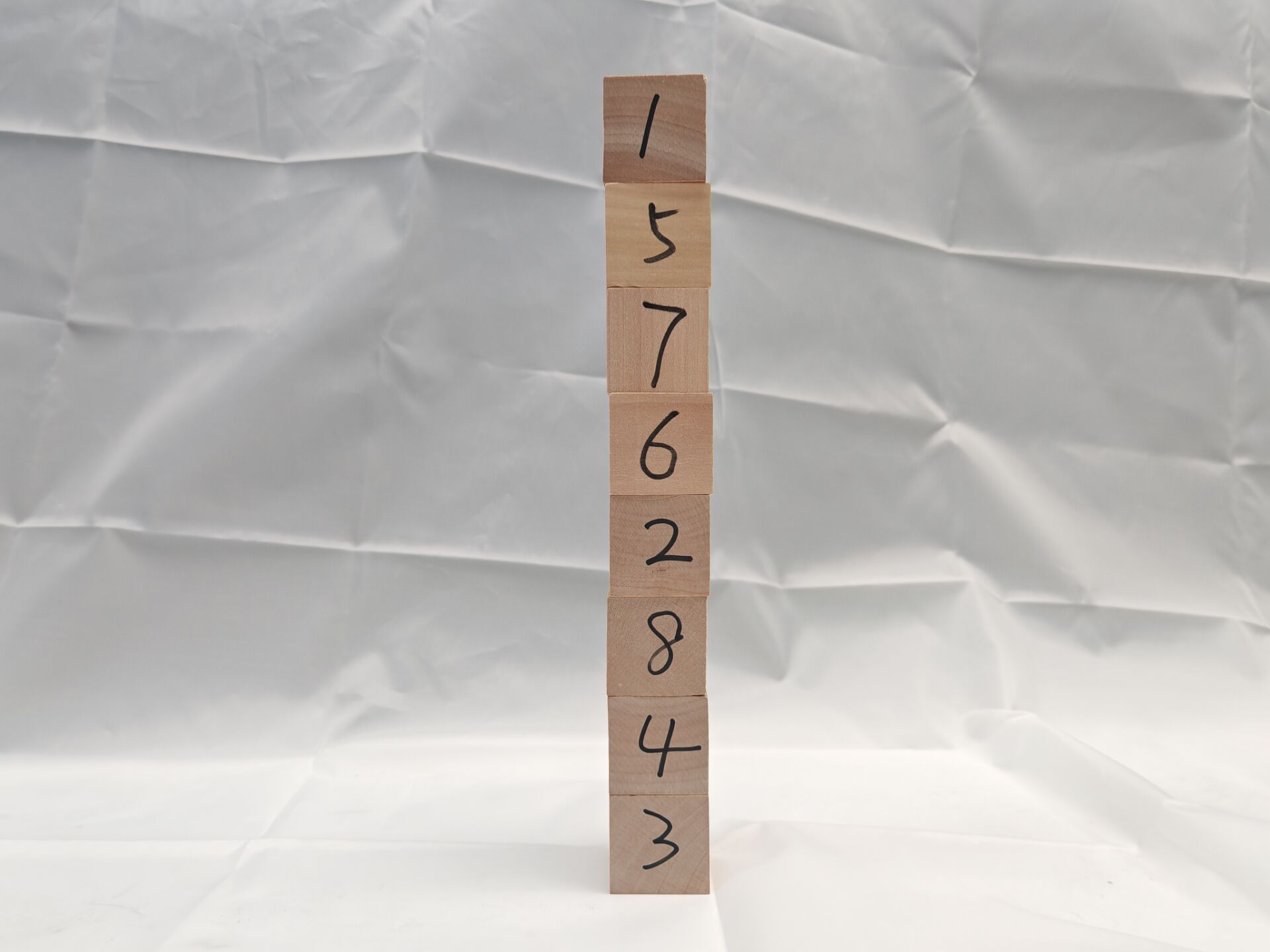}
  \captionof{figure}{Arrange all blocks into two separate stacks on the table. The first stack should have blocks 1, 3, 5, and 7 in order from top to bottom. The second stack should have blocks 2, 4, 6, and 8 in order from top to bottom.}
  \label{fig:blockworld_1}
\end{minipage}\hspace{0.05\textwidth}
\begin{minipage}[t]{0.45\textwidth}
  \includegraphics[width=\linewidth]{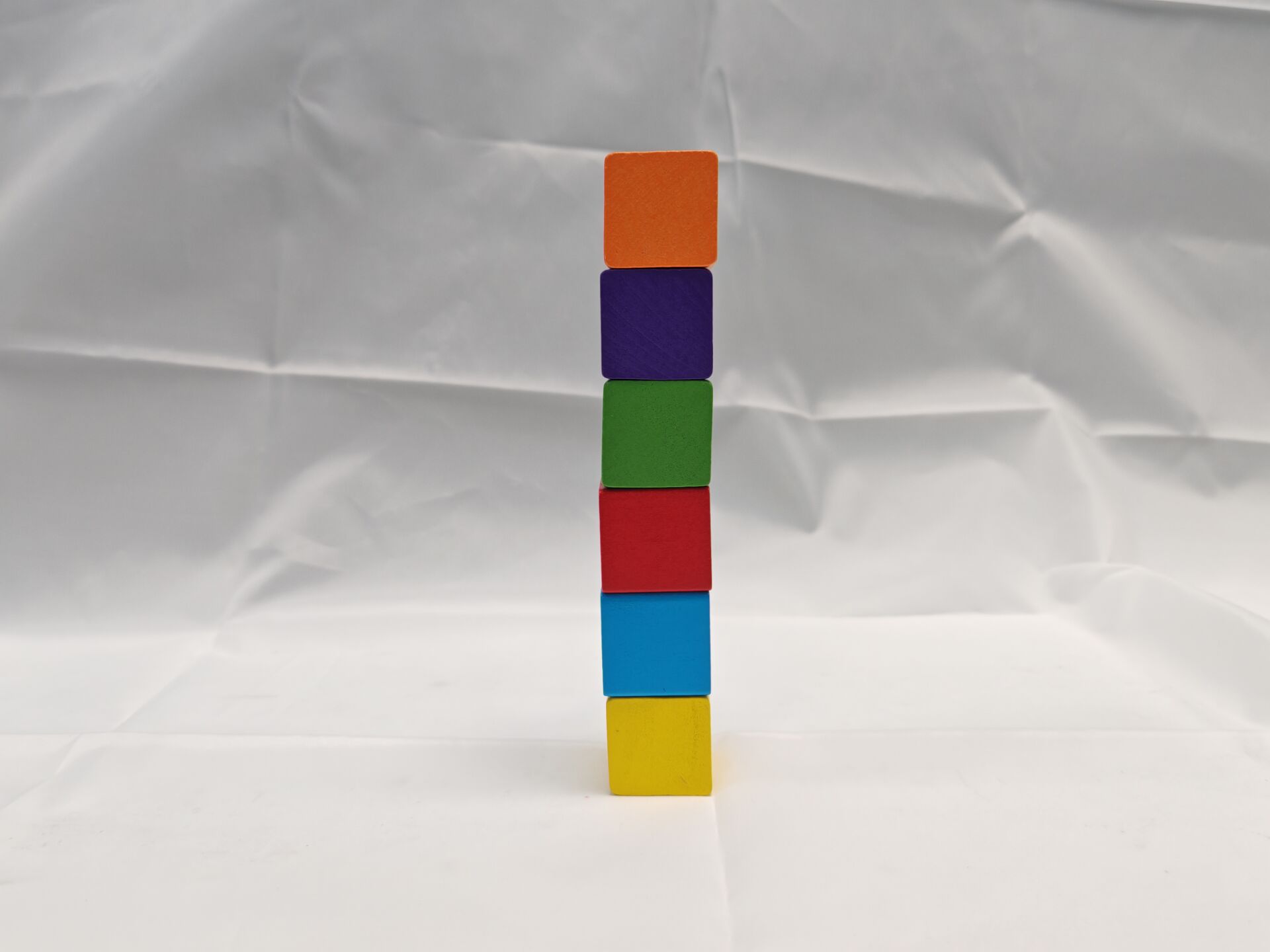}
  \captionof{figure}{Arrange all blocks in a single stack (top to bottom): purple, orange, blue, green, yellow, red.}
  \label{fig:blockworld_2}
\end{minipage}

\vspace{3em}
%----------- Row 2 -----------
\subsection{Representative Tasks in \textit{Desktop}}\label{apx:dsk}

\begin{minipage}[t]{0.45\textwidth}
  \includegraphics[width=\linewidth]{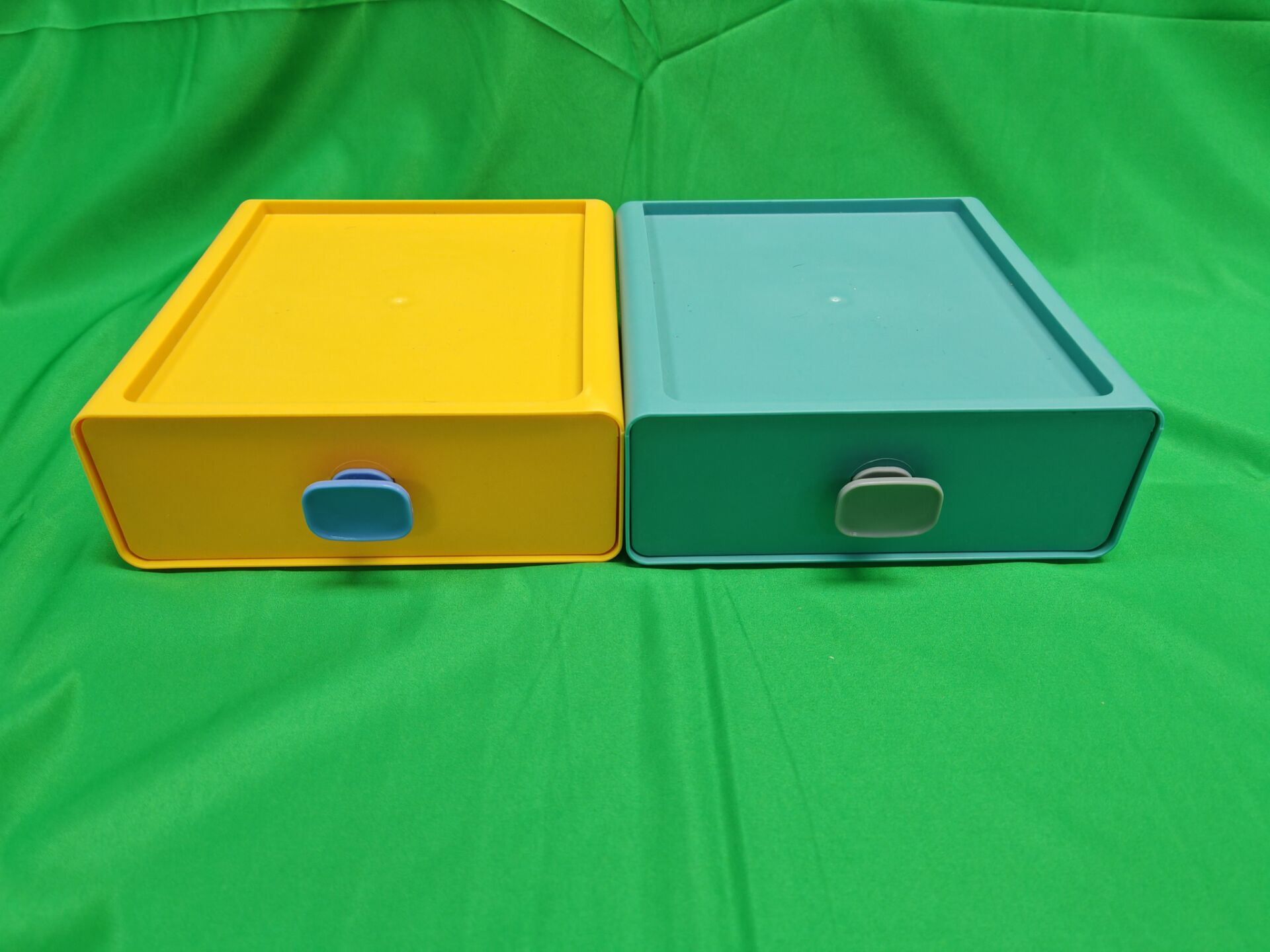}
  \captionof{figure}{There is a block in the green drawer. Please put it on the table, push it and put it in the yellow drawer.}
  \label{fig:desktop_1}
\end{minipage}\hspace{0.05\textwidth}
\begin{minipage}[t]{0.45\textwidth}
  \includegraphics[width=\linewidth]{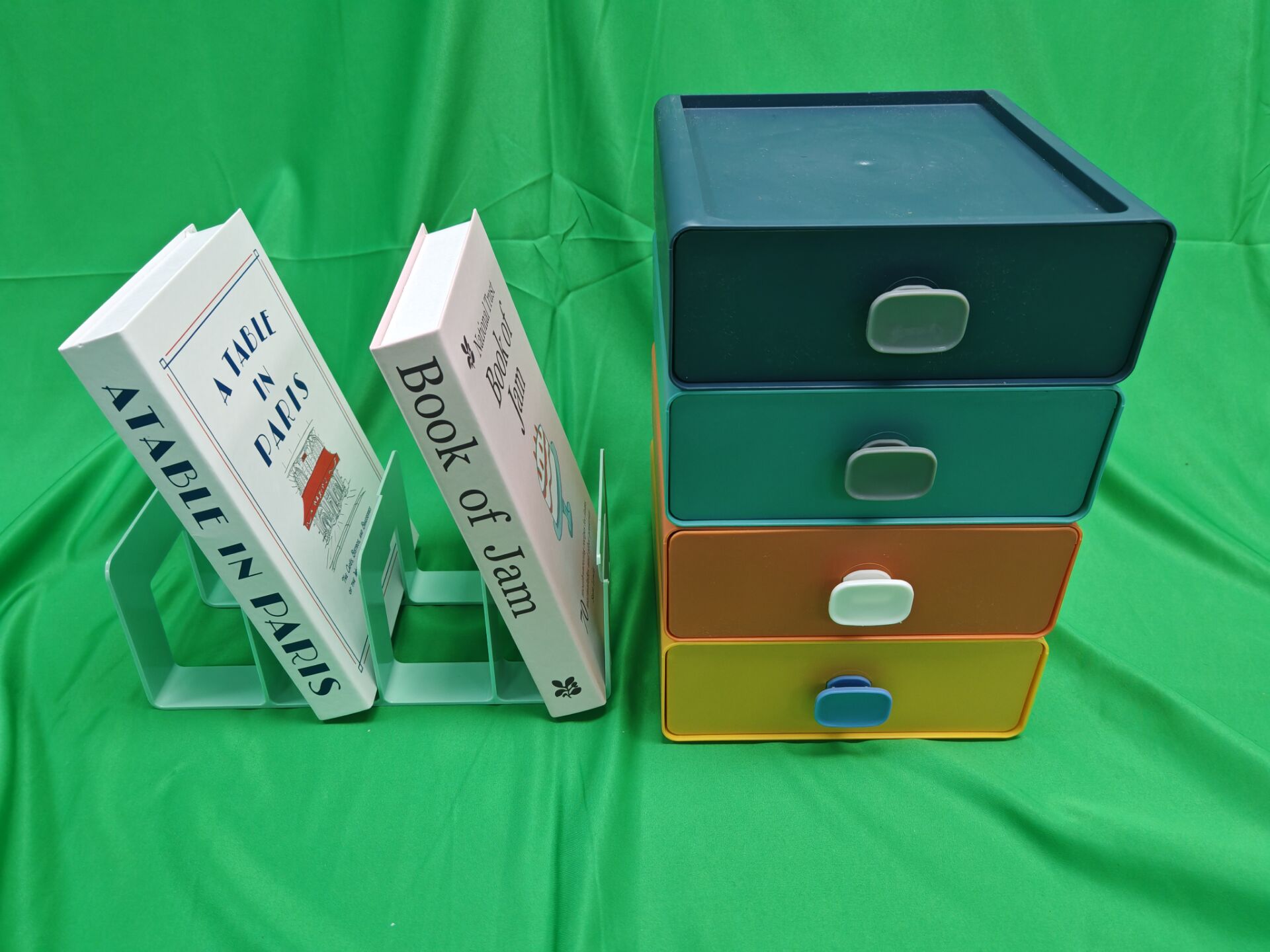}
  \captionof{figure}{There is a tissue in the yellow drawer. Put it on the table and put the white book on the pink book.}
  \label{fig:desktop_2}
\end{minipage}
\end{center}

% \clearpage
%%%%%%%%%%%%%%%%%%%%%%%%%%%%%%%%%%%%%%%%%%%%%%%%%%%%%%%%%%%%%%%%%%%%%%%%
\subsection{Representative Tasks in \textit{Kitchen}}\label{apx:kit}

\begin{center}
%----------- Row 1 -----------
\begin{minipage}[t]{0.45\textwidth}
  \includegraphics[width=\linewidth]{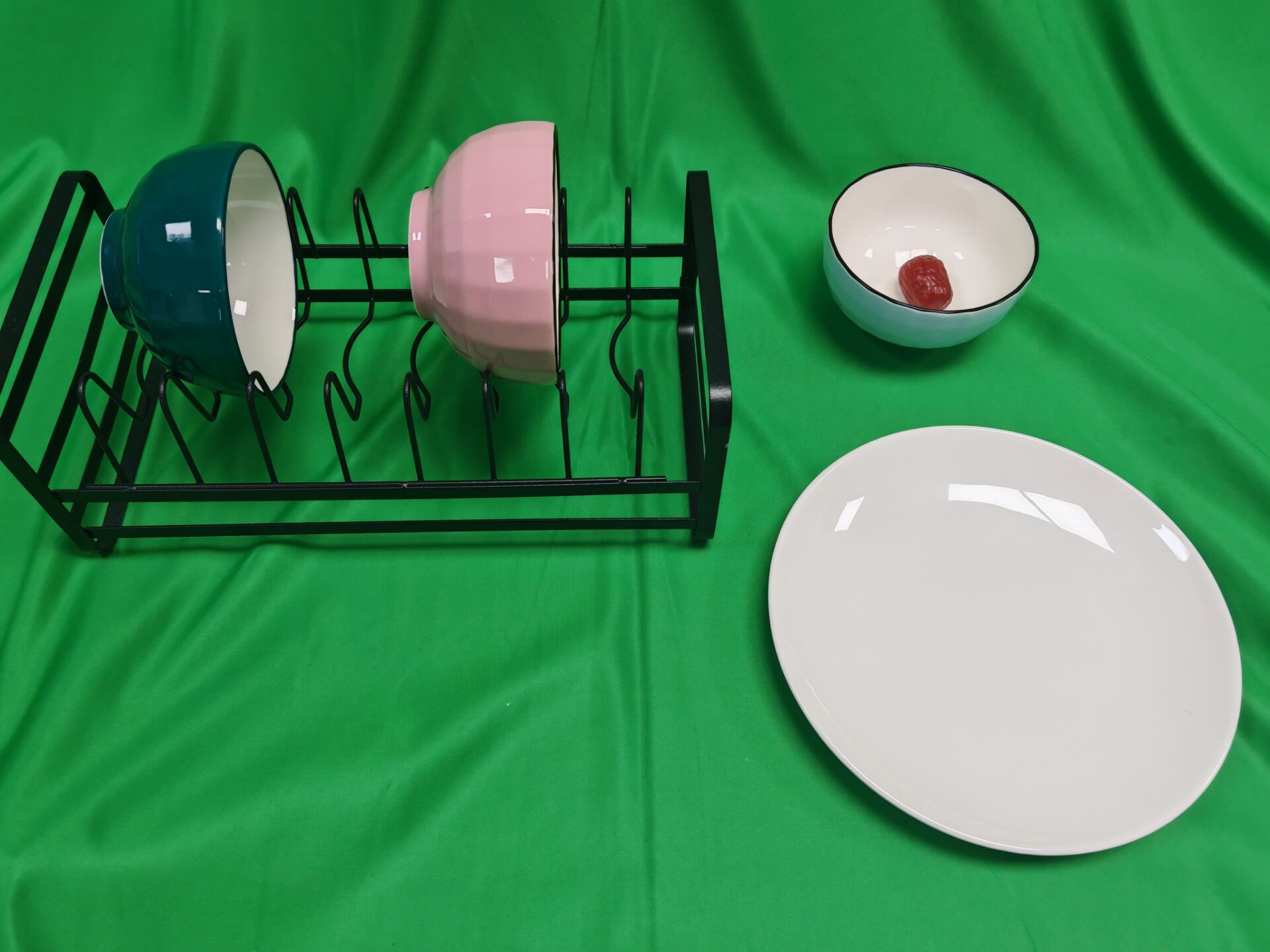}
  \captionof{figure}{Put the jujube in the green bowl. And put the white plate on the rack.}
  \label{fig:kitchen_1}
\end{minipage}\hspace{0.05\textwidth}
\begin{minipage}[t]{0.45\textwidth}
  \includegraphics[width=\linewidth]{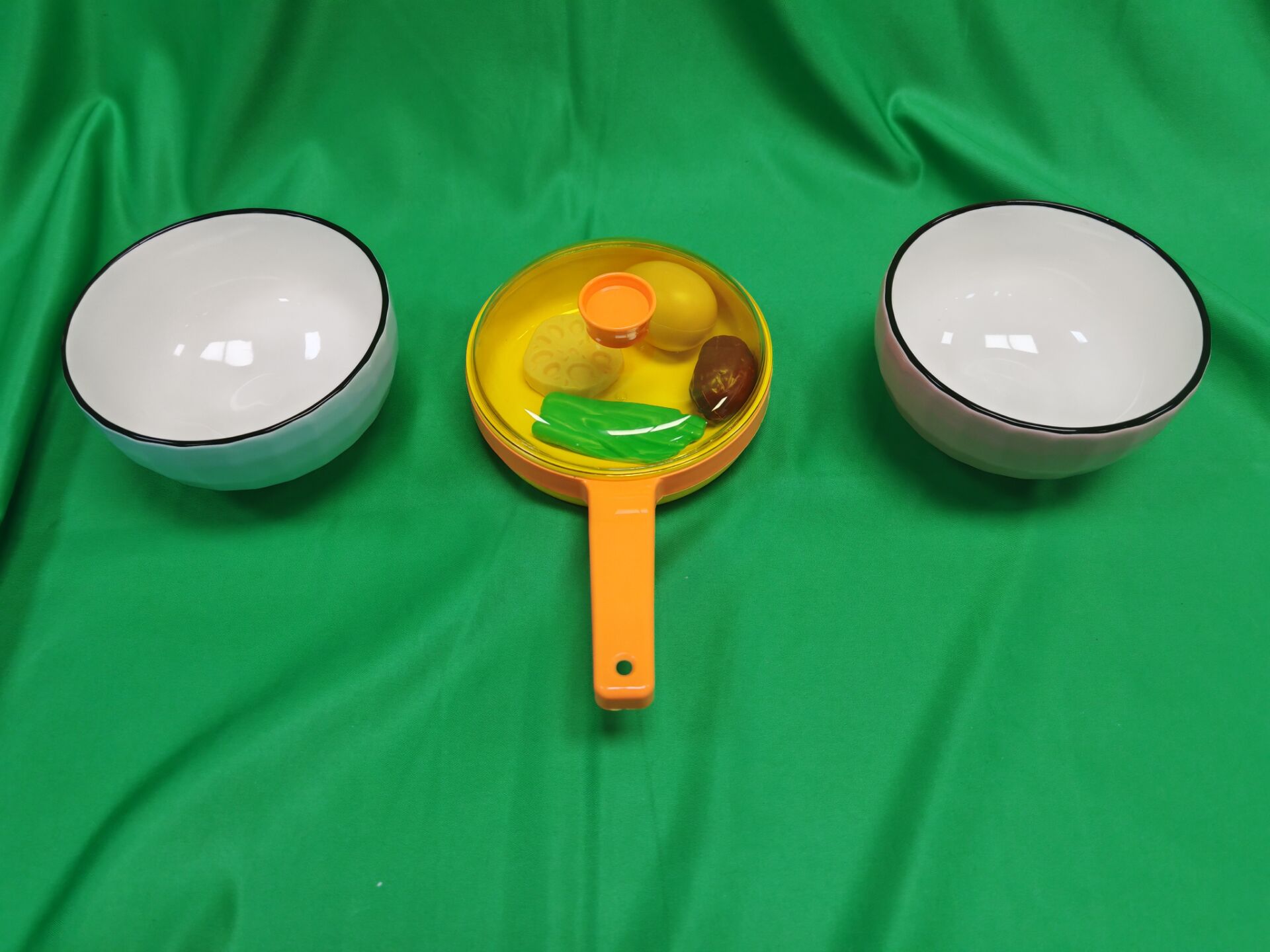}
  \captionof{figure}{Take the egg out of the pot and put it in the left bowl. Take the vegetable out of the pot and put it on the right bowl.}
   \label{fig:kitchen_2}
\end{minipage}
\end{center}
\vspace{3em}
%----------- Row 2 -----------
\subsection{Representative Tasks in \textit{Combination}}\label{apx:com1}

\begin{center}
\begin{minipage}[t]{0.45\textwidth}
  \includegraphics[width=\linewidth]{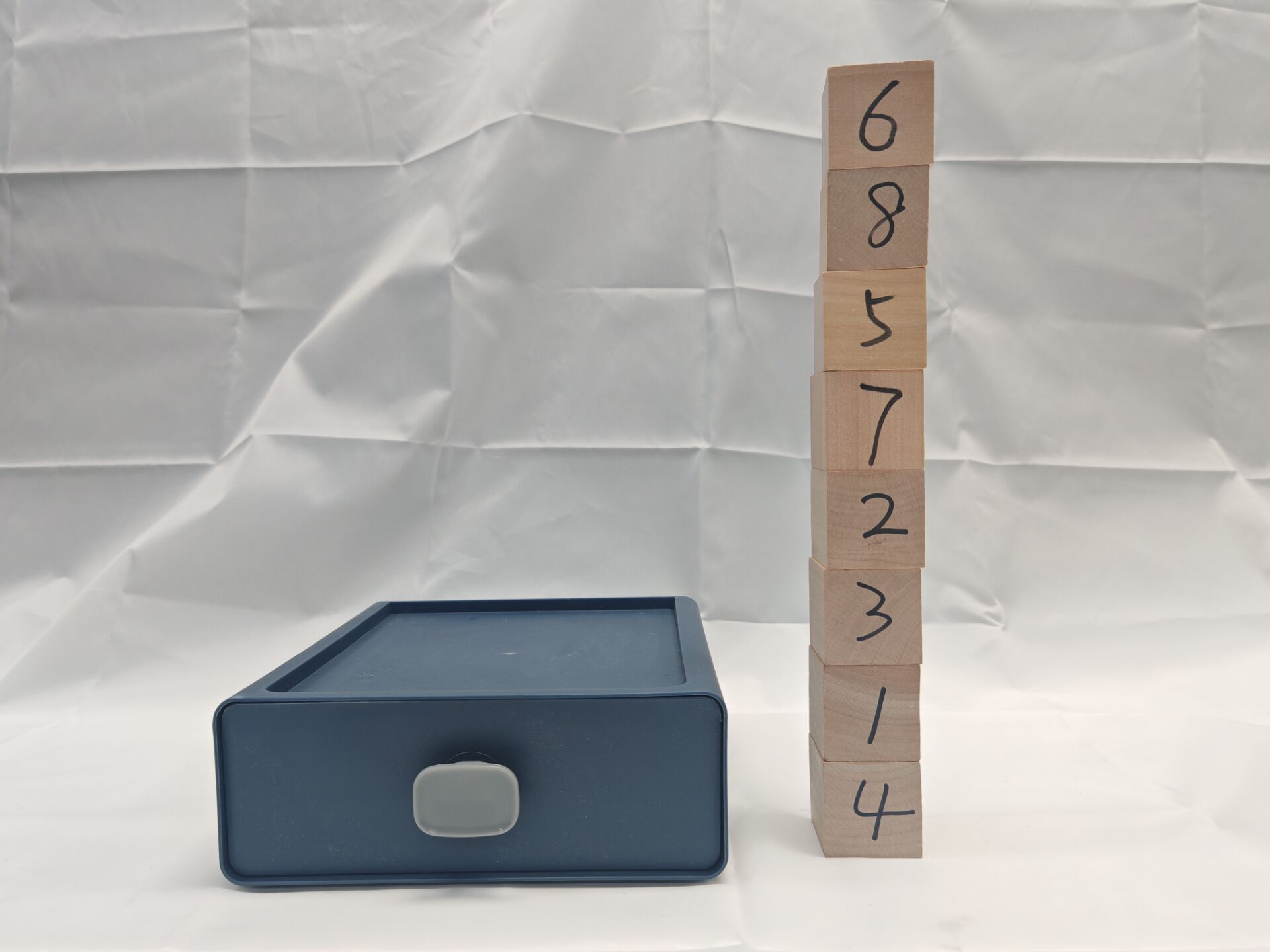}
  \captionof{figure}{Put block 8 in the drawer, and arrange other blocks in a single stack on the table (from top to bottom): 1, 2, 3, 4, 5, 6, 7.}
  \label{fig:combination_1}
\end{minipage}\hspace{0.05\textwidth}
\begin{minipage}[t]{0.45\textwidth}
  \includegraphics[width=\linewidth]{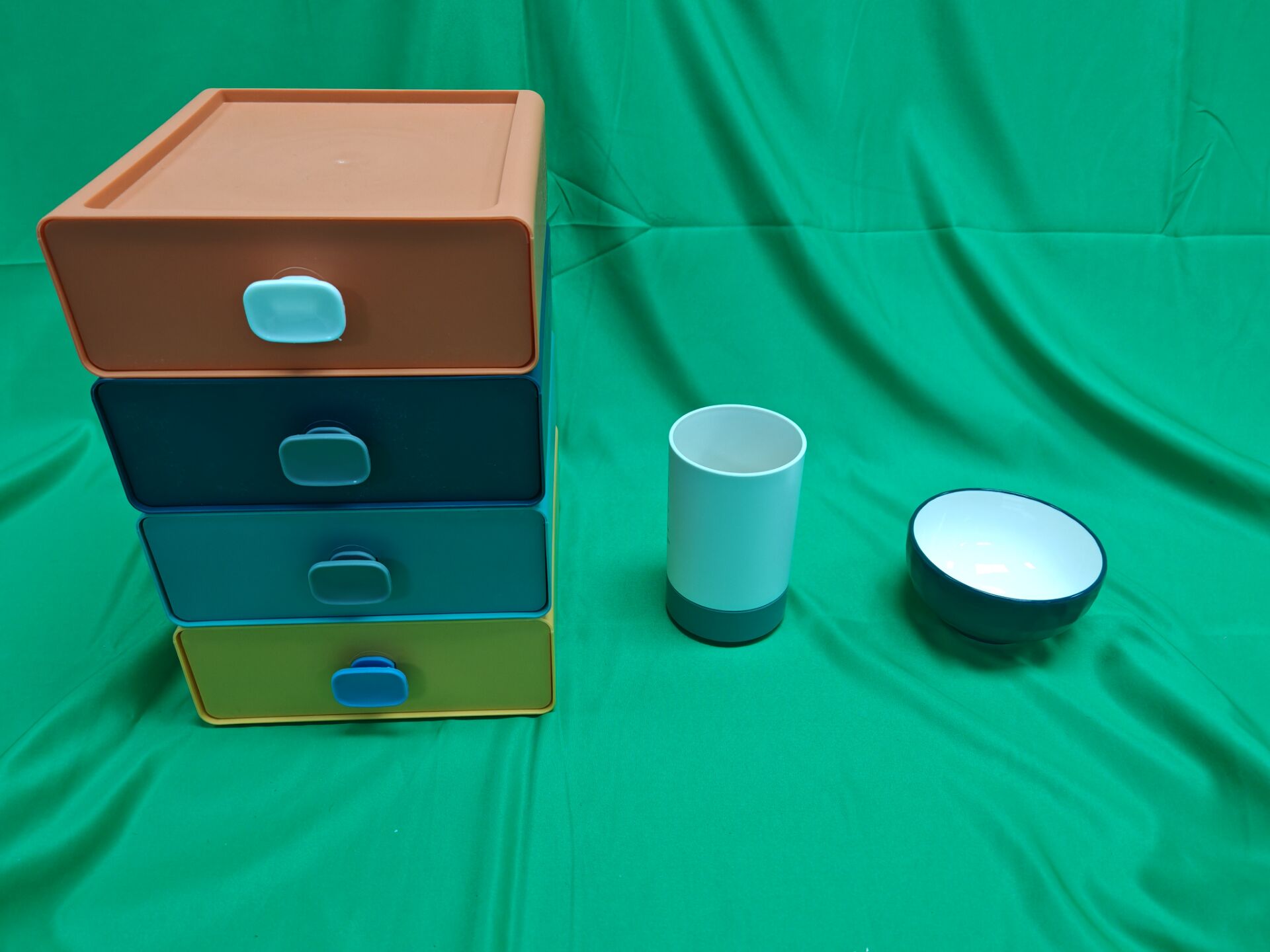}
  \captionof{figure}{There are a spoon, a tissue, an orange block in the green drawer. Stir the bowl and put the spoon in the cup, put the orange block into the orange drawer, wipe the bowl and scrunch the tissue on the table.}
  \label{fig:combination_2}
\end{minipage}
\end{center}

\clearpage

\section{Verb Distribution in the Unified Domain and the Meta Domain}\label{apx:verb}
\begin{figure}[h!]
  \centering

  % 第一行
  \parbox[t]{0.48\textwidth}{
    \centering
    \includegraphics[width=\linewidth]{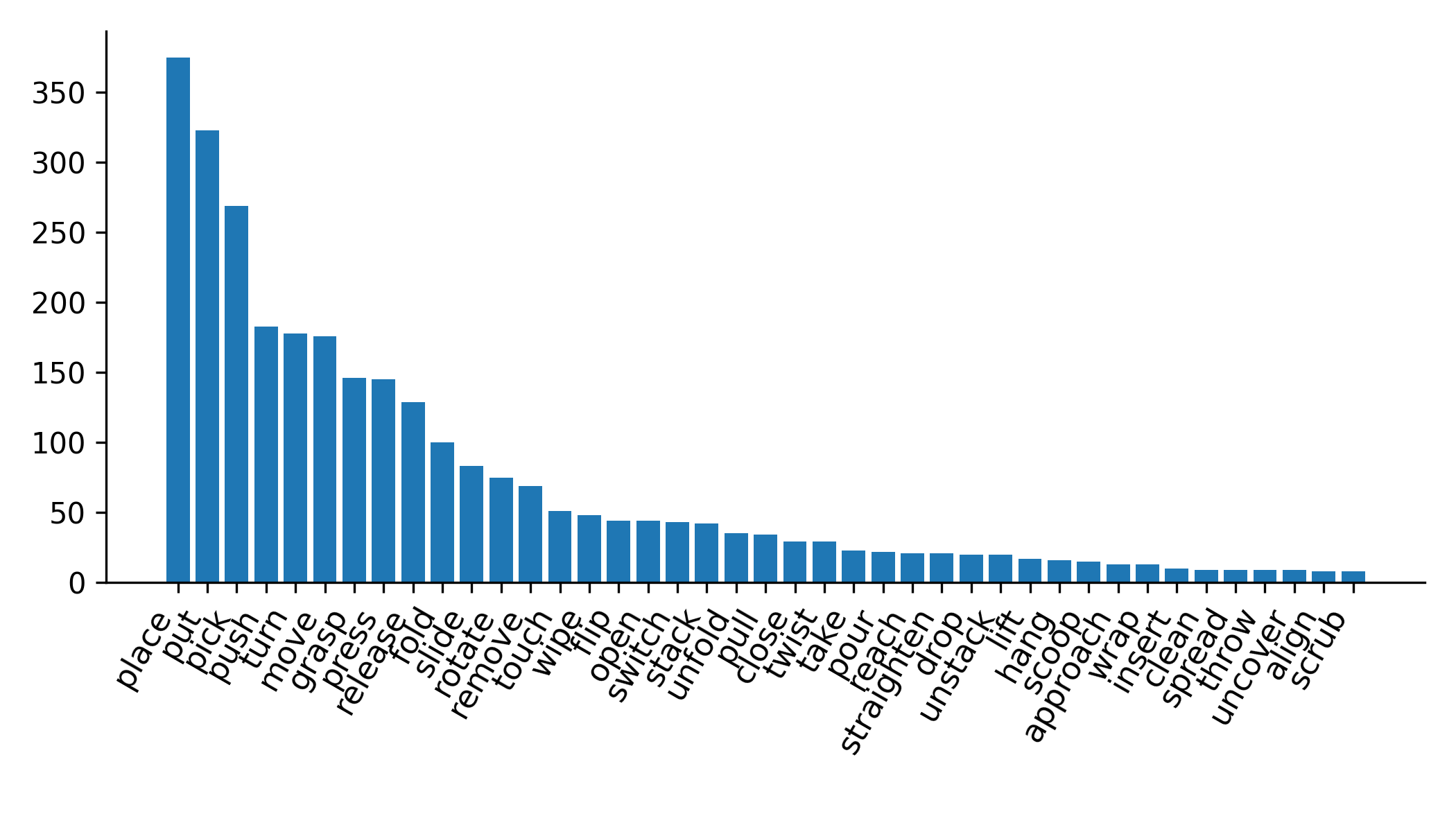}
    \vspace{-0.6em} \\ (a) part 1
  }
  \hfill
  \parbox[t]{0.48\textwidth}{
    \centering
    \includegraphics[width=\linewidth]{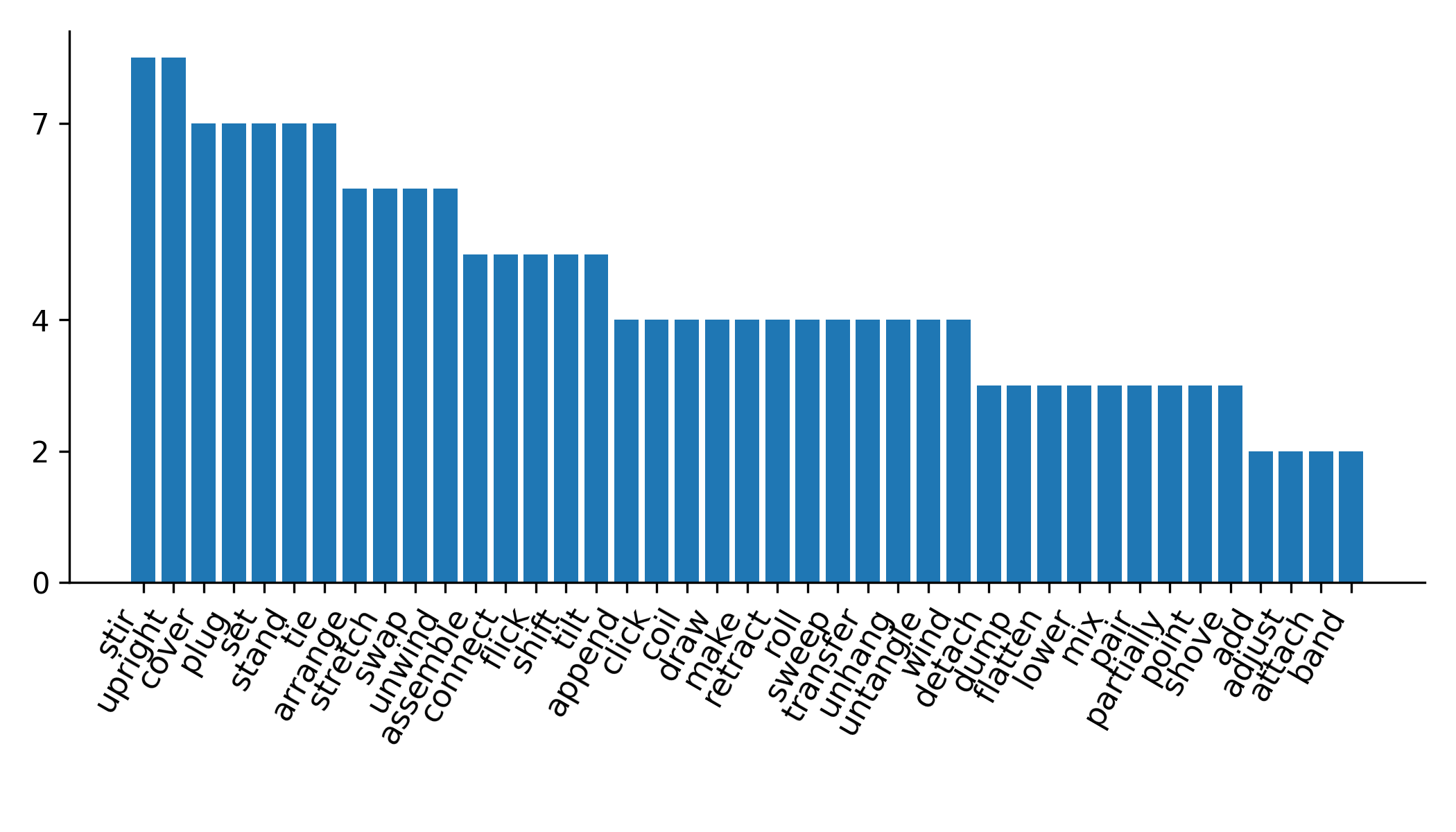}
    \vspace{-0.6em} \\ (b) part 2
  }

  \vspace{1em}

  % 第二行
  \parbox[t]{0.48\textwidth}{
    \centering
    \includegraphics[width=\linewidth]{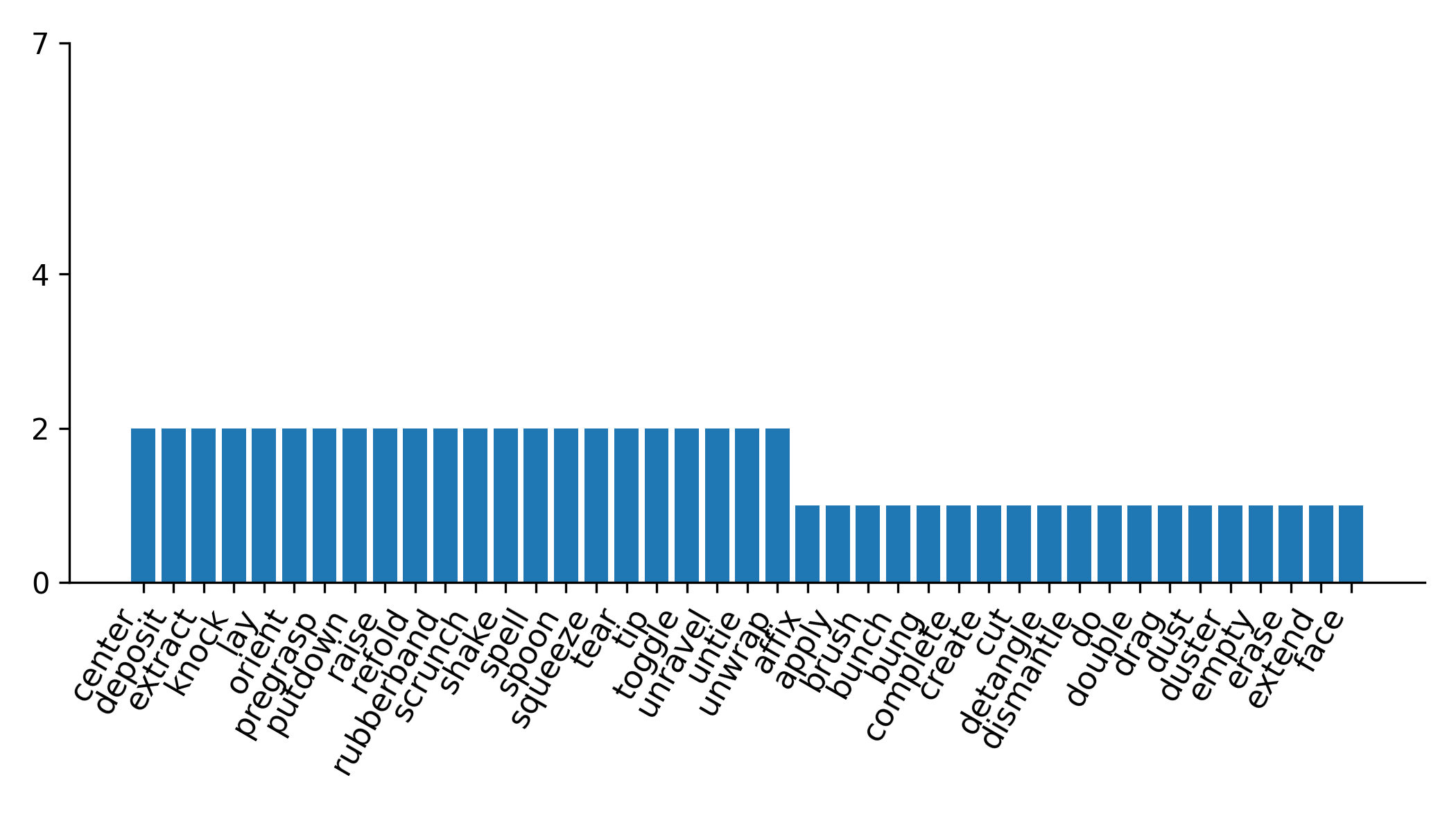}
    \vspace{-0.6em} \\ (c) part 3
  }
  \hfill
  \parbox[t]{0.48\textwidth}{
    \centering
    \includegraphics[width=\linewidth]{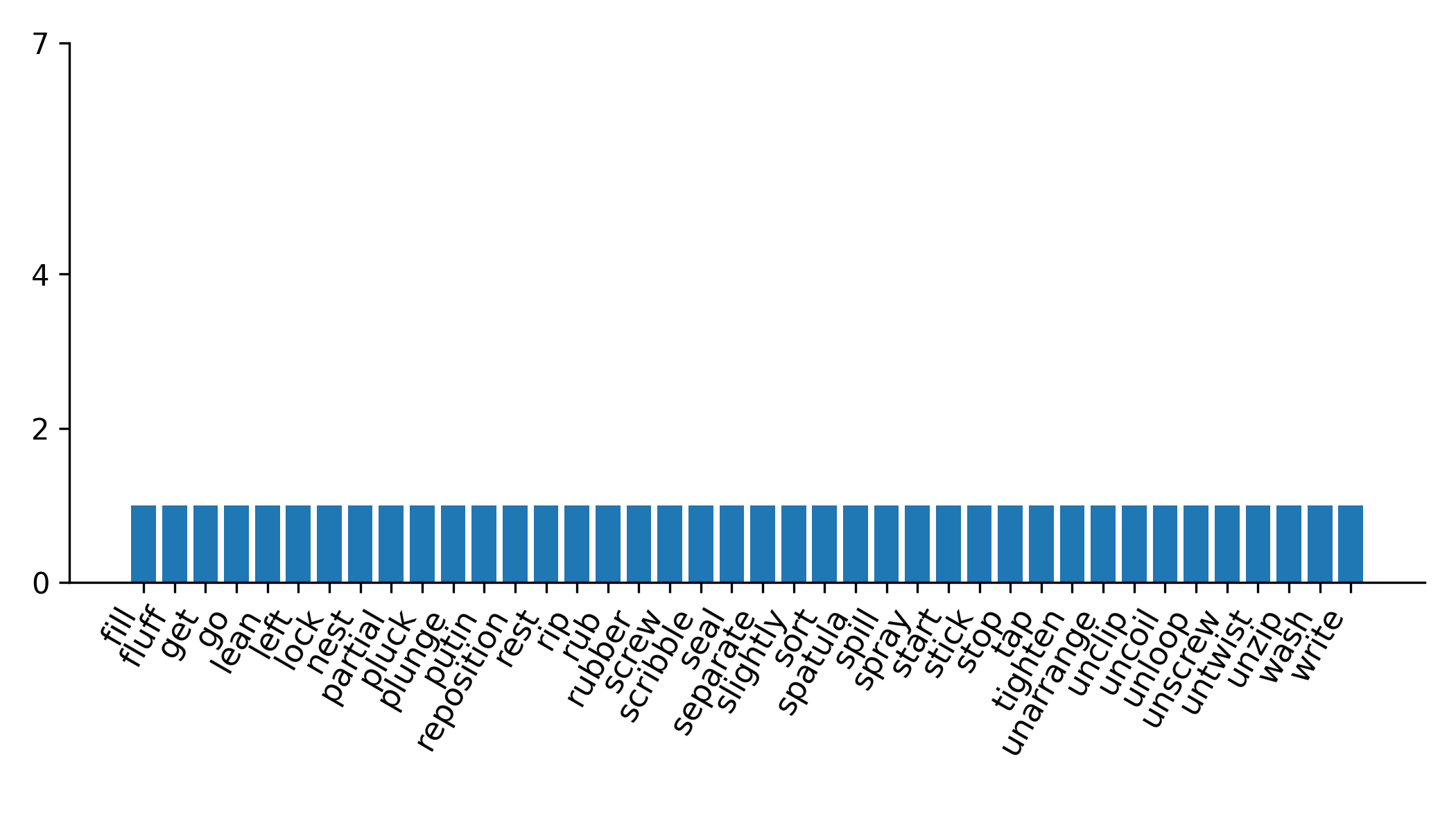}
    \vspace{-0.6em} \\ (d) part 4
  }

  \caption{Verb distribution of our pre-trained unified domain, divided into four segments (a)--(d).}
  \label{fig:action-type-distribution}
\end{figure}

\begin{figure}[b]
    \centering
    \includegraphics[width=0.5\linewidth]{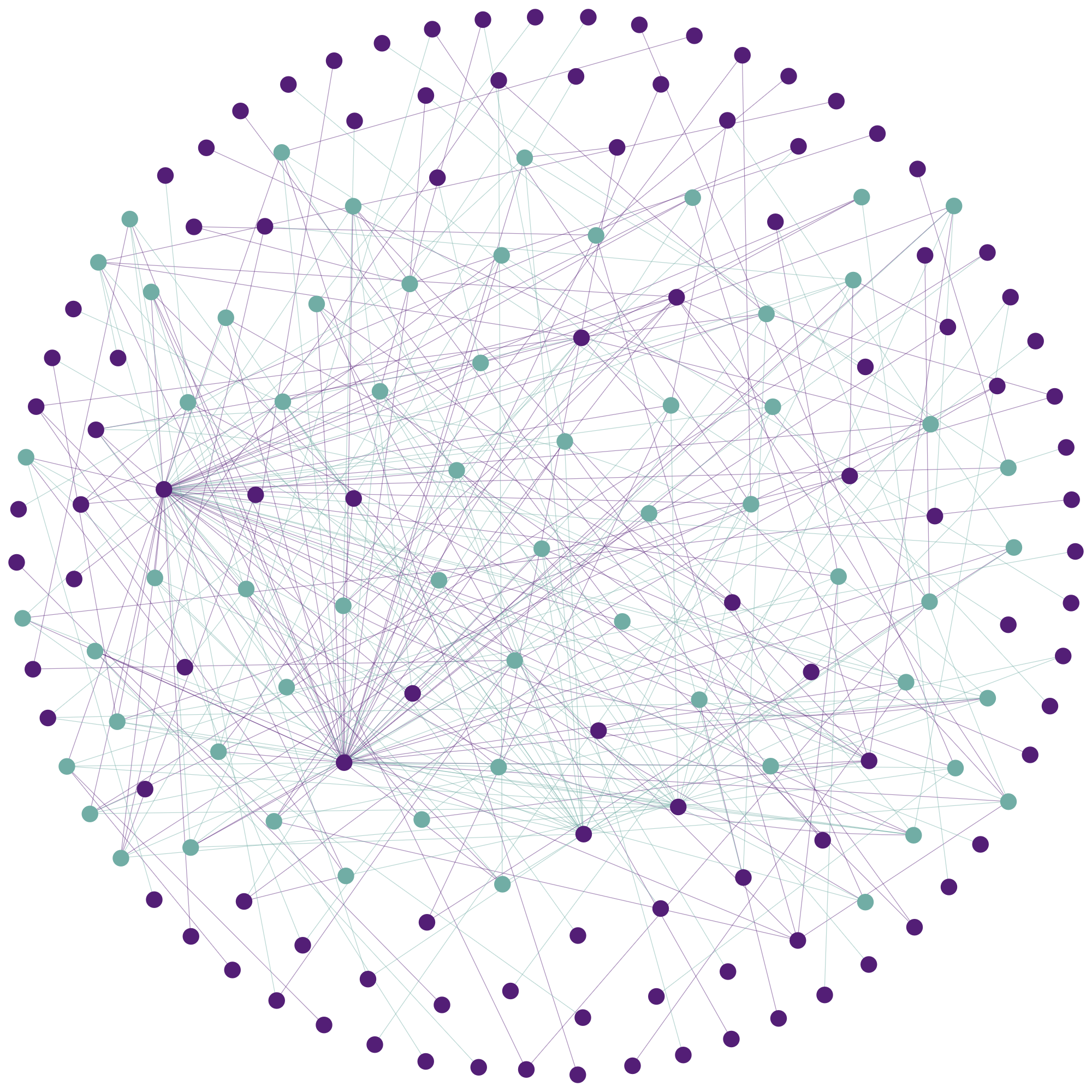}
    \caption{Visualization of meta domain used for planning across evaluation tasks, with 61 operator nodes (green) and 106 predicate nodes (purple).}
    \label{fig:planing_domain}
\end{figure}

\begin{figure}[t]
    \centering
    \includegraphics[width=0.7\linewidth]{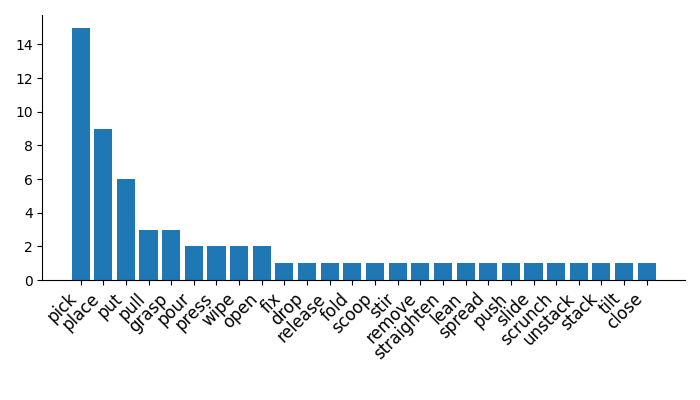}
    \vspace{-0.8em}
    \caption{Verb distribution of meta domain used for planning in evaluation tasks.}
    \label{fig:action type planing domain}
\end{figure}

UniDomain processes 12,393 real-world demonstrations to construct a unified domain consisting of 3,137 operators grouped into 170 verb types (Figure~\ref{fig:action-type-distribution}).
For planning evaluation, we retrieve and fuse 40 household-related atomic domains from this unified set to form a compact meta-domain with 106 predicates (including negated predicates), 61 operators, and 332 causal edges (Figure~\ref{fig:planing_domain}).
This compositional meta domain integrates a diverse set of kitchen-related verbs (Figure~\ref{fig:action type planing domain}).

\newpage
\section{Details and Justification on Energy-based Keyframe Extraction}\label{apx:keyframe}

\textbf{Hyperparameter.} To accommodate videos of varying lengths $l$, we employ an adaptive window size $K$, specified as follows: $K=\begin{cases}
	10,l\leqslant 100\\
	20,100<l\leqslant 150\\
	30,150<l\leqslant 200\\
	40,200<l\leqslant 500\\
	40+10\times \left( \lfloor \frac{l-501}{200} \rfloor + 1\right),l>500\\
\end{cases}$

To explain how and why our seemingly-simple keyframe extraction works, we offer both theoretical and empirical justification below.

\begin{figure}[t]
  \centering
  \includegraphics[width=\linewidth]{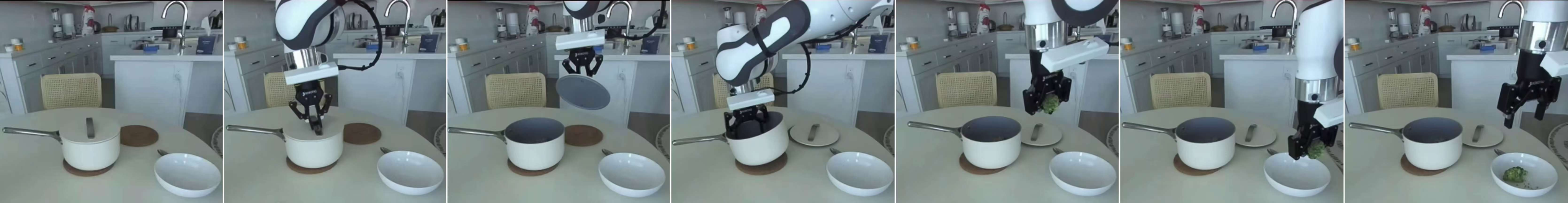}
  \caption{Energy-based keyframes on \textbf{DROID}. The automatically selected frames align with intuitive manipulation boundaries}
  \label{fig:droid_keyframes}
\end{figure}

\textbf{Theoretical Justification.} Our method of summing squared grayscale intensities is equivalent to measuring the total energy of the image treated as a 2D signal. By Parseval's theorem \cite{parseval},  this spatial-domain energy is proportional to the energy in the frequency domain. Semantic transitions in a video, such as an object being picked up or a drawer opening, cause significant changes in the image's structure and texture, which correspond to shifts in its frequency-space energy. Our method identifies keyframes by detecting the local extrema of this energy sequence, effectively capturing these of significant semantic change. Figure ~\ref{fig:droid_keyframes} shows an example set of keyframes extracted from a \textit{DROID} demonstration via our energy-based method. The key-frames successfully captured semantic phase changes (from approach, grasp, lift, to place), echoing our theoretical justification.

\begin{figure}[t]
  \centering
  % 第一个子图 (a)
  \begin{subfigure}[b]{\linewidth}
    \centering
    \includegraphics[width=\linewidth]{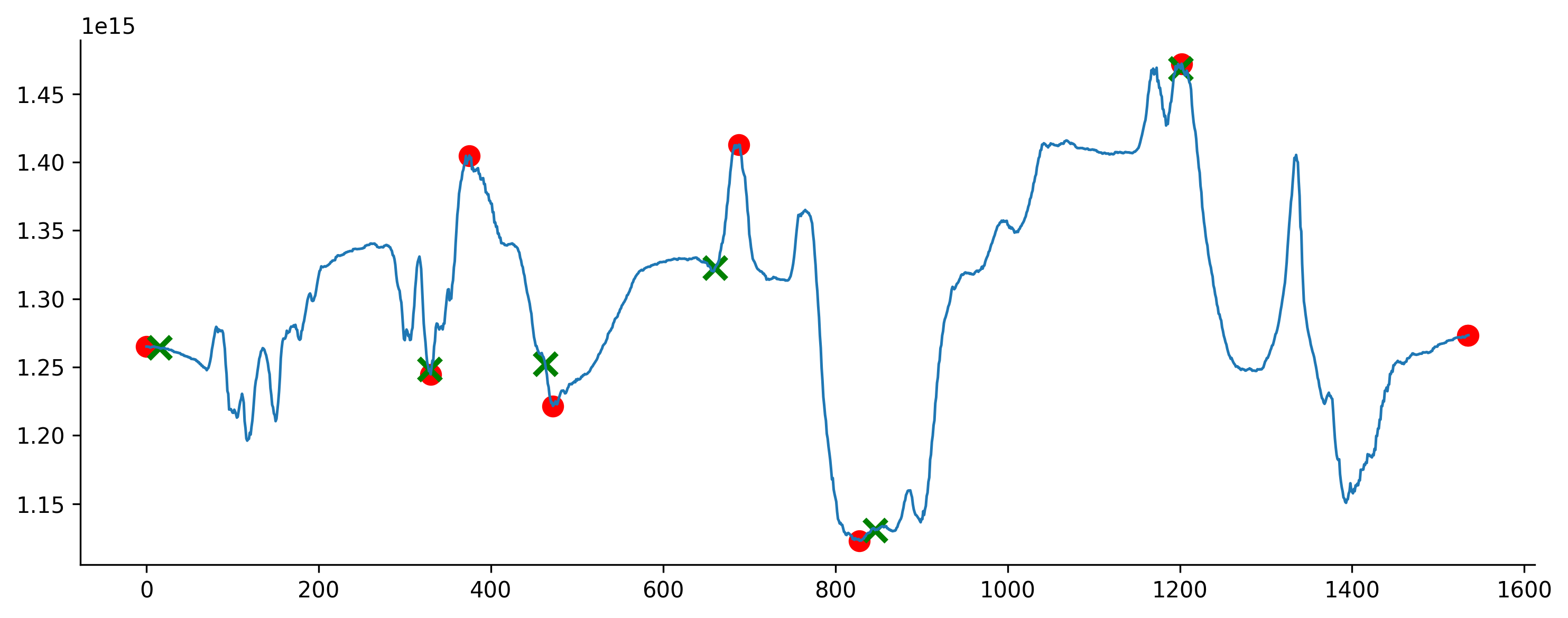}
    \caption{}
    \label{fig:agibot_energy_a}
  \end{subfigure}
  \vspace{1em} 
  % 第二个子图 (b)
  \begin{subfigure}[b]{\linewidth}
    \centering
    \includegraphics[width=\linewidth]{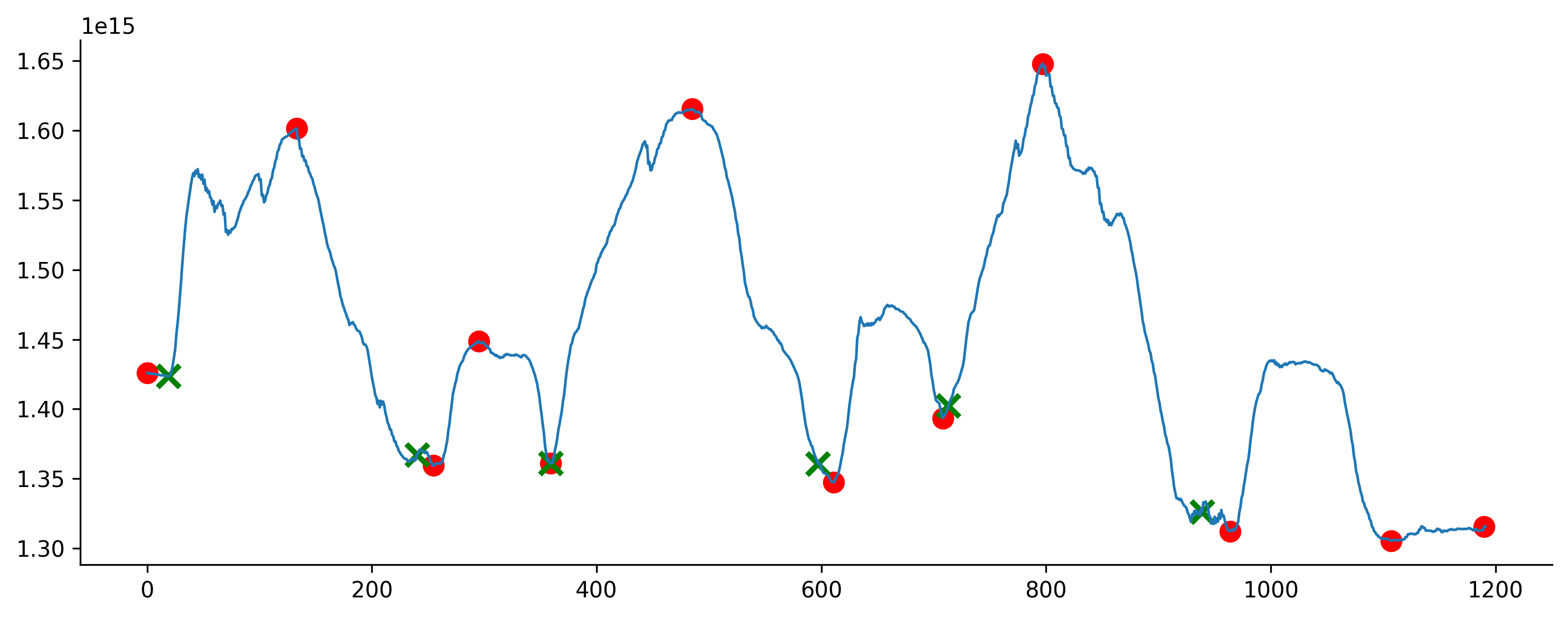}
    \caption{}
    \label{fig:agibot_energy_b}
  \end{subfigure}
  \caption{(a)(b) are two example of the energy curve in Agibot World datasets. Red circles: keyframes extracted by our energy-based method; green crosses: manual annotations in Agibot World. The horizontal axis represents the frame index, and the vertical axis represents the energy value.}
  \label{fig:agibot_energy}
\end{figure}

\textbf{Empirical Justification.} We use the Agibot World \cite{agibot} dataset, which provides human-annotated keyframes along with demonstrations, for empirical justification. We compare our automatically-extracted keyframes with human-annotated keyframes provided by Agibot World. Energy curves shown in Figure \ref{fig:agibot_energy_a} and Figure \ref{fig:agibot_energy_b} show that human-annotated keyframes consistently cluster around the local energy extrema we extracted, confirming the effectiveness of our method. 

\textbf{Connection to Primitive Actions.} "Primitive actions", or "primitive skills", in a common terminology used in task planning (or task and motion planning) research \cite{primitive1, primitive2, primitive3}. A primitive skill is an atomic, specialized skill for manipulating one object or one object–relation tuple, such as "pick an object" or "open a container", inducing atomic, local effects, and acting as a building block for longer tasks. Our energy-based keyframe extraction detects local extrema in frame-energy curves, either a peak or a trough, that consistently occurs at primitive skill boundaries (for example, when a move ends and a pick begins). When spliting the demonstration at these extrema, each segment would span one primitive skill, resulting in the learned operator being atomic.

% \begin{figure}[t]
%   \centering
%   \includegraphics[width=\linewidth]{images/503_719648_energy_plot_100.png}
%   \caption{One example of the energy curve in Agibot World datasets. Red circles: energy-based keyframes; green crosses: manual annotations. Manual picks concentrate near energy extrema.}
%   \label{fig:agibot_energy_a}
% representative failure casesrepresentative failure cases% \end{figure}

% \begin{figure}[t]
%   \centering
%   \includegraphics[width=\linewidth]{images/385_680075_energy_plot_80.png}
%   \caption{Another example of the energy curve in Agibot World datasets. Red circles: energy-based keyframes; green crosses: manual annotations. Manual picks concentrate near energy extrema.}
%   \label{fig:agibot_energy_b}
% \end{figure}

\section{Evaluation Results and Failure Examples Per Task Class}\label{apx:eval}
\begin{figure}[h]
    \centering

    % 图例
    \includegraphics[width=0.9\linewidth]{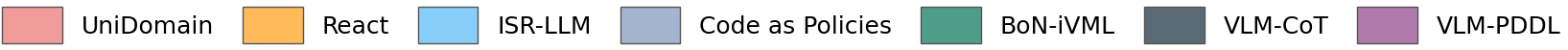} \\
    \vspace{0.6em}

    % 第一行: SR和SPL，居中且紧凑
    \begin{minipage}{0.47\linewidth}
        \centering
        \includegraphics[width=0.93\linewidth]{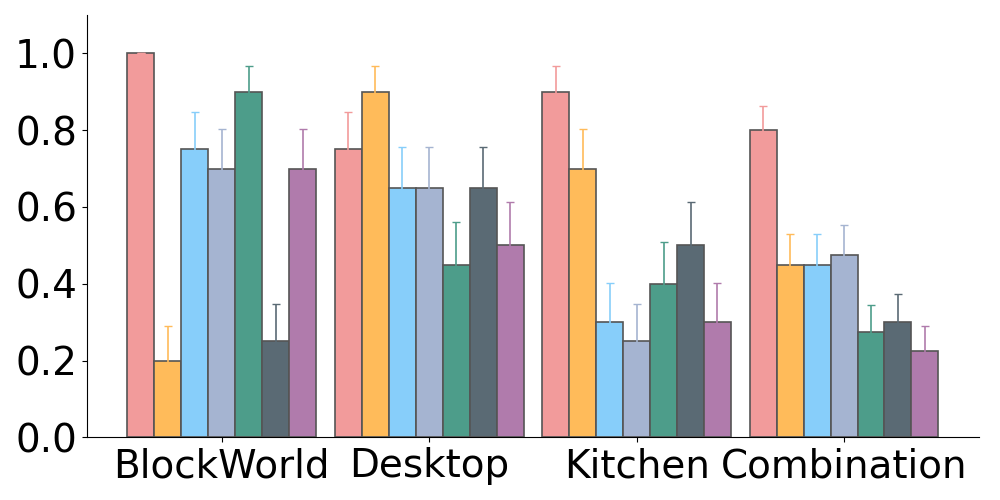}
        \\ \vspace{-0.2em} (a) SR
    \end{minipage}
    \hspace{0.04\linewidth} % 控制两图间距
    \begin{minipage}{0.47\linewidth}
        \centering
        \includegraphics[width=0.93\linewidth]{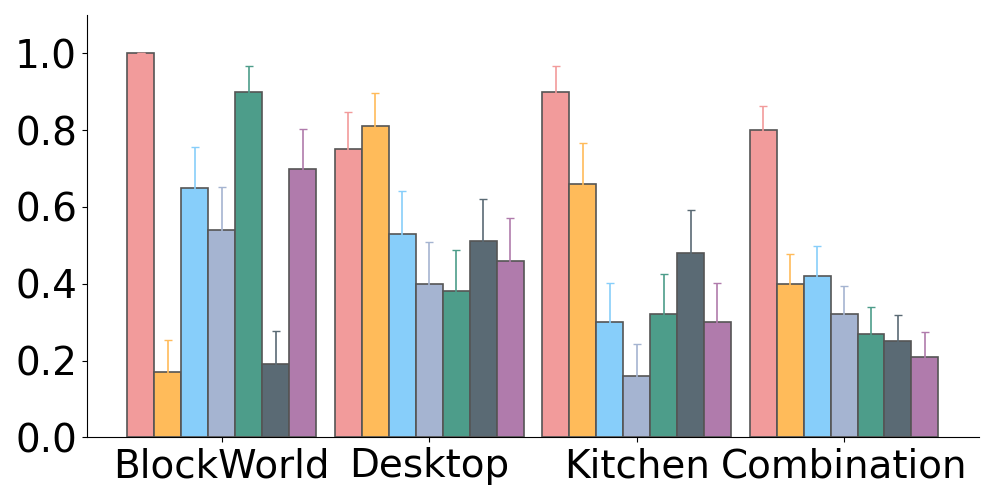}
        \\ \vspace{-0.2em} (b) SPL
    \end{minipage}

    \vspace{1.0em}

    % 第二行: OR2, OR1, OR0
    \begin{minipage}{0.32\linewidth}
        \centering
        \includegraphics[width=0.95\linewidth]{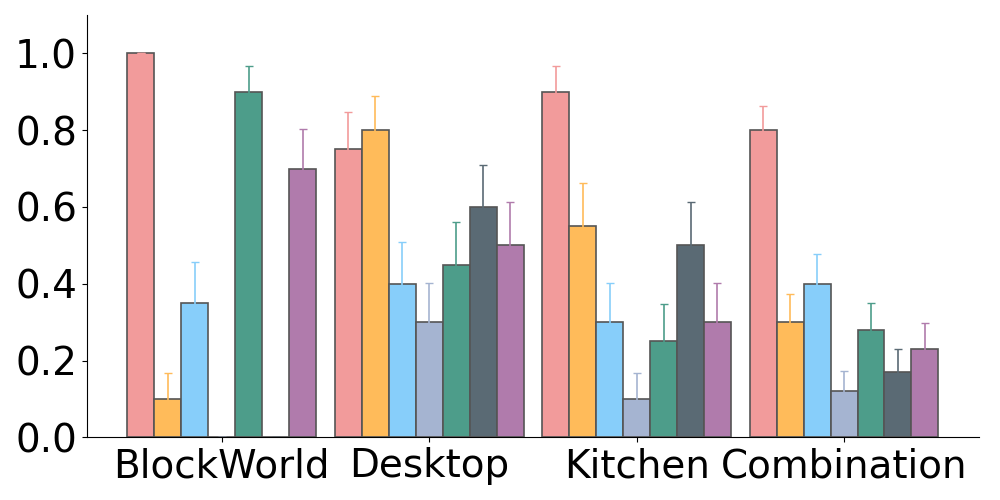}
        \\ \vspace{-0.2em} (c) OR (K=2)
    \end{minipage}
    \begin{minipage}{0.32\linewidth}
        \centering
        \includegraphics[width=0.95\linewidth]{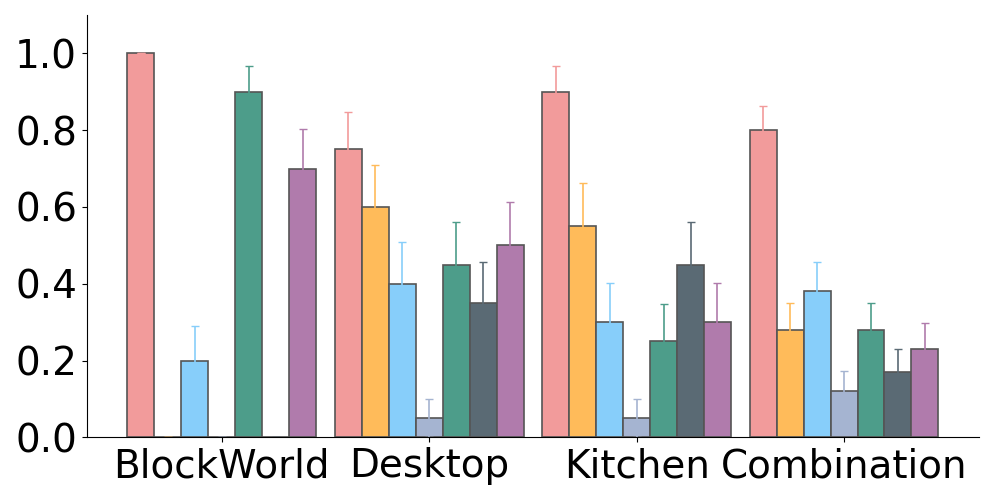}
        \\ \vspace{-0.2em} (d) OR (K=1)
    \end{minipage}
    \begin{minipage}{0.32\linewidth}
        \centering
        \includegraphics[width=0.95\linewidth]{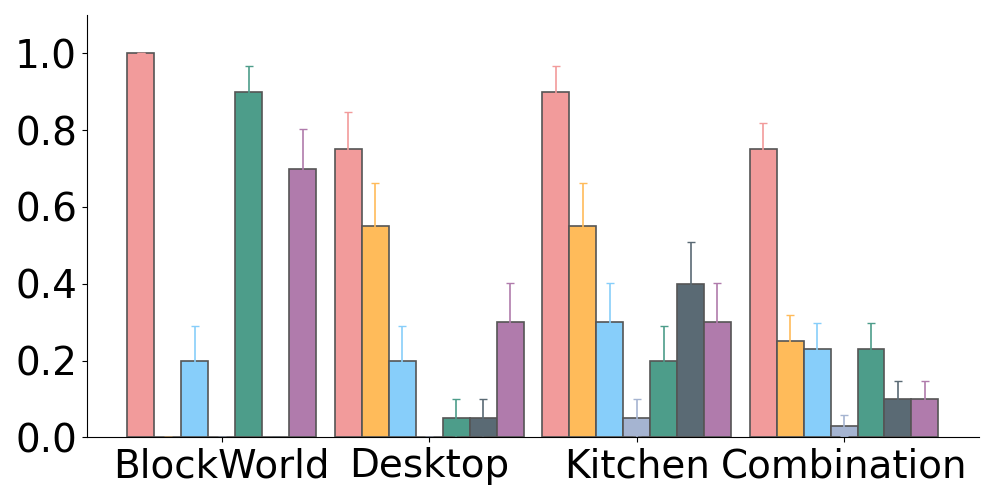}
        \\ \vspace{-0.2em} (e) OR (K=0)
    \end{minipage}

    \caption{Separate comparison results of UniDomain and state-of-the-art baselines on the four classes of real-world tasks. Average values are shown with standard errors.}
    \label{fig:main_results}
\end{figure}

Figure \ref{fig:main_results} reports comparison results in the four evaluation task classes separately, also evaluating three key metrics: Success Rate (SR), Success weighted by Path Length (SPL), and Optimality Rate (OR) at increasing strictness levels (K=2,1,0). Below, we provide a qualitative comparison between the best-performing and the worst-performing methods in the four task classes.

% This metric progression systematically transitions evaluation focus from task feasibility to plan optimality. 
% All methods were assessed using GPT-4.1 API (temperature=0.0) under identical environment setups.

% In what follows, we substantiate the aggregate success rate by comparing the strongest method and the weakest baseline within per task class on representative instances, thereby explaining the factors behind the superior performance of UniDomain.

\paragraph{BlockWorld.}
% In~\autoref{fig:blockworld_1}, the language instruction is ``\textit{Arrange all blocks into two separate stacks on the table. The first stack should have blocks 1, 3, 5, and 7 in order from top to bottom. The second stack should have blocks 2, 4, 6, and 8 in order from top to bottom}'', and the worst-performing method (React) attempted to perform an infeasible action, commanding to pick block 3 without first unstacking block 4 on top of it. The best-performing method (UniDomain) correctly recognized this physical constraint.

In~\autoref{fig:blockworld_1}, the worst-performing method (React) attempted to perform an infeasible action, commanding to pick block 3 without first unstacking block 4 on top of it. The best-performing method (UniDomain) correctly recognized this physical constraint.

\noindent\textit{Task.}
\lstset{basicstyle=\ttfamily\small, showspaces=false}
\begin{lstlisting}
Arrange all blocks into two separate stacks on the table. The first
stack should have blocks 1, 3, 5, and 7 in order from top to bottom.
The second stack should have blocks 2, 4, 6, and 8 in order from top
to bottom.
\end{lstlisting}

\noindent\textit{React plan (failed).}
\begin{lstlisting}[basicstyle=\ttfamily\small]
pick up block 1 from the top of the stack
put down block 1 on the table to start the first stack
pick up block 5 from the top of the stack
put down block 5 on top of block 1 
pick up block 7 from the top of the tall stack
put down block 7 on top of block 5 
pick up block 3 from the bottom of the second stack
pick up block 6 from the top of the second stack
put down block 6 on the table temporarily
pick up block 2 from the top of the second stack
put down block 2 on the table 
pick up block 6 from the table
put down block 6 on the table 
pick up block 2 from the table
put down block 2 on top of block 4 
pick up block 6 from the table
pick up block 2 from the table
pick up block 3 from the bottom of the second stack
pick up block 7 from the top of the first stack
pick up block 7 from the top of the first stack
pick up block 6 from the table
pick up block 6 from the table
pick up block 6 from the table
\end{lstlisting}

\noindent\textit{UniDomain PDDL problem (showing only core predicates).}
\begin{lstlisting}[basicstyle=\ttfamily\small]
(:init
    (clear block1)
    (on block1 block5)
    (on block5 block7)
    (on block7 block6)
    (on block6 block2)
    (on block2 block8)
    (on block8 block4)
    (on block4 block3)
    (on_table block3 table)
    (hand_free robot)
)
(:goal 
  (and 
    (clear block1) 
    (on block1 block3) 
    (on block3 block5) 
    (on block5 block7) 
    (on_table block7 table)
    (clear block2)
    (on block2 block4) 
    (on block4 block6) 
    (on block6 block8) 
    (on_table block8 table) 
  )
)
\end{lstlisting}

\noindent\textit{UniDomain plan (succeeded).}
\begin{lstlisting}[basicstyle=\ttfamily\small]
(unstack robot block1 block5)
(place_on_table robot block1 table)
(unstack robot block5 block7)
(place_on_table robot block5 table)
(unstack robot block7 block6)
(place_on_table robot block7 table)
(pick_from_table robot block5 table)
(stack robot block5 block7)
(unstack robot block6 block2)
(stack robot block6 block5)
(unstack robot block2 block8)
(place_on_table robot block2 table)
(unstack robot block8 block4)
(place_on_table robot block8 table)
(unstack robot block6 block5)
(stack robot block6 block8)
(unstack robot block4 block3)
(stack robot block4 block6)
(pick_from_table robot block2 table)
(stack robot block2 block4)
(pick_from_table robot block3 table)
(stack robot block3 block5)
(pick_from_table robot block1 table)
(stack robot block1 block3)

\end{lstlisting}

\paragraph{Desktop.}
In~\autoref{fig:desktop_1}, the worst-performing method (BoN-iVML) commanded to open the wrong drawer, distinct from the one instructed, and omitted a critical ``push'' action. The best-performing method (React) correctly understood these key task elements.

\noindent\textit{Task.}
\begin{lstlisting}[basicstyle=\ttfamily\small, breaklines=true, breakindent=0pt]
There is a block in the green drawer. Please put it on the table, push it and put it in the yellow drawer.
\end{lstlisting}

\noindent\textit{BoN-iVML PDDL problem.}
\begin{lstlisting}[basicstyle=\ttfamily\small]
(:init
    (block_in_green_drawer block1)
    (not (drawer_open green_drawer))
    (not (drawer_open yellow_drawer))
    (robot_handempty)
)
(:goal 
    (block_in_yellow_drawer block1)
)
\end{lstlisting}

\newpage
\noindent\textit{BoN-iVML plan (failed).}
\begin{lstlisting}[basicstyle=\ttfamily\small]
(open_drawer yellow_drawer)
(pick_up_block_from_green_drawer block1 yellow_drawer)
(put_down_block_in_yellow_drawer block1 yellow_drawer)
\end{lstlisting}

\noindent\textit{React plan (succeeded).}
\begin{lstlisting}[basicstyle=\ttfamily\small]
open the green drawer
pick up the block from the green drawer
put down the block on the table
push the block on the table
open the yellow drawer
pick up the block from the table
put down the block in the yellow drawer
\end{lstlisting}

\paragraph{Kitchen.}
In~\autoref{fig:kitchen_1}, the worst-performing method (Code as Policies) attempted to place an object into a bowl that is still leaning on a vertical drying rack. The highest-performing method (UniDomain) correctly captured this operational commonsense.

\noindent\textit{Task.}
\begin{lstlisting}[basicstyle=\ttfamily\small]
Put the jujube in the green bowl. And put the white plate on the rack.
\end{lstlisting}

\noindent\textit{Code as Policies plan (failed).}
\begin{lstlisting}[basicstyle=\ttfamily\small]
pick_up(jujube)
put_down(green bowl)
pick_up(white plate)
put_down(rack)
\end{lstlisting}

\noindent\textit{UniDomain PDDL problem (showing only core predicates).}
\begin{lstlisting}[basicstyle=\ttfamily\small]
(:init
    (on_rack bowl_pink rack1)
    (on_rack bowl_green rack1)
    (on_table bowl_white table1)
    (on_table plate_white table1)
    (in_bowl jujube1 bowl_white)
    (hand_free robot1)
)
(:goal 
  (and 
    (in_bowl jujube1 bowl_green) 
    (on_rack plate_white rack1)
  )
)
\end{lstlisting}

\noindent\textit{UniDomain plan (succeeded).}
\begin{lstlisting}[basicstyle=\ttfamily\small]
(pick_from_rack robot1 bowl_green rack1)
(place_on_table robot1 bowl_green table1)
(pick_from_bowl robot1 jujube1 bowl_white)
(put_in_bowl robot1 jujube1 bowl_green table1)
(pick_from_table robot1 plate_white table1)
(place_on_rack robot1 plate_white rack1)
\end{lstlisting}

\paragraph{Combination.}
In~\autoref{fig:combination_2}, the worst-performing method (VLM-PDDL) suffered from syntax-errors and incomplete operators in its PDDL problems and domains. The best-performing method (UniDomain) ensured the quality of the PDDL domain and the syntax correctness of PDDL problems.

\newpage
\noindent\textit{Task.}
\lstset{basicstyle=\ttfamily\small, showspaces=false}
\begin{lstlisting}
There are a spoon, a tissue, an orange block in the green drawer. Stir 
the bowl and put the spoon in the cup, put the orange block into the 
orange drawer, wipe the bowl and scrunch the tissue on the table.
\end{lstlisting}

\noindent\textit{VLM-PDDL PDDL domain (part).}
\begin{lstlisting}[basicstyle=\ttfamily\small]
(:action open
    :parameters (?dr - drawer)
    :precondition (drawer-closed ?dr)
    :effect (and (drawer-open ?dr) (not (drawer-closed ?dr)))
)
(:action pick-up
    :parameters (?obj - object ?loc - container)
    :precondition (and (at ?obj ?loc) (not (in-hand ?any - object)))
    :effect (and (in-hand ?obj) (not (at ?obj ?loc)))
)
\end{lstlisting}

\noindent\textit{UniDomain PDDL domain (part).}
\begin{lstlisting}[basicstyle=\ttfamily\small]
(:action open_drawer 
     :parameters (?r ?x) 
     :precondition (and (hand_free ?r) (drawer ?x) (not (is_open ?x))) 
     :effect (and (is_open ?x))
)
(:action pick_from_drawer
    :parameters (?r ?o ?d)
    :precondition(and(hand_free?r)(in_drawer?o?d)(is_open?d)(drawer?d))
    :effect(and(not(in_drawer?o?d))(not(hand_free ?r))(holding ?r ?o))
)
\end{lstlisting}

\noindent\textit{UniDomain PDDL problem (showing only core predicates).}
\begin{lstlisting}[basicstyle=\ttfamily\small]
(:init
    (in_drawer orange_block1 drawer_green)
    (in_drawer spoon1 drawer_green)
    (in_drawer tissue1 drawer_green)
    (on_table bowl1 table1)
    (on_table cup1 table1)
    (unfolded tissue1)
    (can_wipe_table tissue1)
    (can_wipe_bowl tissue1)
    (can_stir_bowl spoon1)
    (hand_free robot1)
)
(:goal 
  (and 
    (stirred bowl1)
    (wiped bowl1)
    (scrunched tissue1)  
    (in_cup spoon1 cup1) 
    (in_drawer orange_block1 drawer_orange) 
    (on_table tissue1 table1)
  )
)
\end{lstlisting}

\newpage
\noindent\textit{UniDomain plan (succeeded).}
\begin{lstlisting}[basicstyle=\ttfamily\small]
(open_drawer robot1 drawer_green)
(pick_from_drawer robot1 spoon1 drawer_green)
(stir_bowl robot1 spoon1 bowl1)
(place_in_cup robot1 spoon1 cup1)
(pick_from_drawer robot1 tissue1 drawer_green)
(wipe_bowl robot1 tissue1 bowl1)
(scrunch_on_table robot1 tissue1 table1)
(open_drawer robot1 drawer_orange)
(pick_from_drawer robot1 orange_block1 drawer_green)
(place_in_drawer robot1 orange_block1 drawer_orange)
\end{lstlisting}

\section{UniDomain Main Failure Modes}\label{apx:failure}

In this section, we present the main failure modes of UniDomain.

\textbf{Domain learning failures} arise from:
% imperfect domain induction and were quantitatively analyzed during our closed-loop domain generation process. These include:  
(1) Syntax errors, e.g., ill-formed parentheses, account for 48\% of failures;  
(2) Missing operators---there are critical actions omitted in 39\% of failed domains;  
(3) Logical conflicts---contradictory preconditions and effects exist within an operator, constituting 35\% of failures;  
(4) Invalid tests---the generated test problems contain invalid or unreachable goal states in 26\% of failures.  

% The above numbers sum over 100\% due to overlapping error modes in individual failures. 
Crucially, our closed-loop refinement always successfully detects and corrects all the above issues; hence, the success rate of the closed-loop pipeline is 100\%.

\textbf{Online planning failures} are mostly caused by perception and grounding issues:  
(1) Goal misinterpretation---incorrect interpretation of task goals (e.g., reversed stacking order), occurred in 13.3\% of failures;  
(2) Visual grounding errors---mislocalized or misclassified objects, leading to incorrect initial states, occurred in 53.4\% of failures;  
(3) Insufficient domain coverage---the retrieved meta-domain lacks necessary operators or predicates to support the test task, which occurred in 33.3\% of failures.

% \textcolor{red}{Overall, most planning errors stem from structural deficiencies in the initially learned domains, while visual grounding inaccuracies constitute the second major cause. Incorporating schema-consistency checks, invariant validation, and perception-to-symbol verification significantly enhances plan executability and stability.}

\newpage
\section{Prompt Design of the UniDomain Planner}\label{prompt}
To facilitate symbolic planning with the meta-domain, UniDomain employs a prompt to guide two crucial modules in the online planning: predicate/operator filtering and problem generation. This prompt is used twice, first to identify task-relevant predicates and operators by analyzing the initial problem proposal, and second to refine the final problem file based on the filtered symbolic space. 
\begin{lstlisting}[basicstyle=\ttfamily\small, breaklines=true, breakindent=0pt]
Now you need to help a robot to generate a PDDL problem file based on the given PDDL domain.
The given image shows the initial scene.
The robot's hand is free initially (even though it is not shown in the image) and PDDL objects must include a robot.
The task is under table-top environment and PDDL objects must include at least a table.

Instructions: {instructions}
given PDDL domain: {domain}

Your output should include four parts:
(1) reasoning: analyze the image and output the reasons.
(2) objects: you need to locate objects related to the task from the iamge.
(3) init: you need to describe PDDL init from the image based on given PDDL predicates.
(4) goal: you need to gnerate PDDL goal from human instructions.

When generating PDDL init, you should use type predicate, state predicates, spatial or position relationship predicates and affordance predicates in order.

Your output should be in JSON format like below:
{{
    "reasoning:": "your analysis",
    "objects": ["apple_1", "apple_2", "bowl"],
    "init": ["(on_table bowl)"],
    "goal": "(and (in apple_1, bowl) (in apple_2, bowl))"
}}
\end{lstlisting}

\newpage
\section{Setting and Prompts for Baselines}
\subsection{Baseline Setting}
In our evaluation, all LLM-based methods were adapted to use VLMs, to accept scene image as input and support visual planning. For ReAct, we provide online visual feedback and action executability (success/failure only) to perform closed-loop planning. The maximum closed-loop action steps (including both execution and thinking steps) are constrained to twice the optimal plan cost. For BoN-iVML, we make two modifications in our implementation: (1) employing multi-LLM voting for best-of-N selection, and (2) adding a PDDL problem file generation module, which uses VLM to ensure visual grounding.

\subsection{Prompt for VLM-CoT.}
\begin{lstlisting}[basicstyle=\ttfamily\small, breaklines=true, breakindent=0pt]
Now you need to help a single-armed robot to plan to finish the given task.
The given image shows the initial scene.

The available action APIs for the robot are:
(1) pick up: pick up some object.
(2) put down: put down the object in the robot's hand on/in some object.
(3) open: open some object such as drawer, door, etc.
(4) close: close some object such as drawer, door, etc.
(5) fold: fold some object such as towel, tissue, etc.
(6) wipe: wipe some object using the object in the robot's hand.
(7) scrunch: scrunch some object such as tissue, etc
(8) stir: stir some object using the object in the robot's hand.
(9) push: push some object such as block, etc.
(10) slide: slide some object such as block, etc.
(11) press: press some object such as button, etc.
(12) turn on: turn on some object such as light, tap, etc.
(13) turn off: turn off some object such as light, tap, etc
(14) pull: pull some object such as rod, handle, etc.
(15) pour: pour some object using the object in the robot's hand.
(16) lean: lean some object against some other object, such as lean a board against a wall.

When you make decisions, you should consider constraints based on your common sense.
For example, the robot cannot pick up another object when it is already holding one because it is single-armed.

Instructions: {instructions}

Your output should include two parts:
(1) reasoning: analyze the image and output the reasons.
(2) plan_sequence: output the sequence of actions in natural language to complete the task.

Your output should be in JSON format like below (Do NOT output comments):
{{
    "reasoning:": "your analysis",
    "plan_sequence": ["pick up the green block from table", "place the green block on the red block"]
}}
\end{lstlisting}

\newpage
\subsection{Prompt for Code as policies.}
\begin{lstlisting}[basicstyle=\ttfamily\small, breaklines=true, breakindent=0pt]
    You are an expert at writing modular Python functions for a single-armed robot to complete a task using a fixed set of atomic APIs and the image.
    The given image shows the initial scene. Observe all objects and their spatial relationships, then write code based on this initial state to complete the task.

    The available action APIs for the robot are:
    (1) pick_up(obj): pick up some object.
    (2) put_down(target): put down the object in the robot's hand on/in some object.
    (3) open(obj): open some object such as drawer, door, etc.
    (4) close(obj): close some object such as drawer, door, etc.
    (5) fold(obj): fold some object such as towel, tissue, etc.
    (6) wipe(obj): wipe some object using the object in the robot's hand.
    (7) scrunch(obj): scrunch some object such as tissue, etc
    (8) stir(obj): stir some object using the object in the robot's hand.
    (9) push(obj): push some object such as block, etc.
    (10) slide(obj): slide some object such as block, etc.
    (11) press(obj): press some object such as button, etc.
    (12) turn on(obj): turn on some object such as light, tap, etc.
    (13) turn off(obj): turn off some object such as light, tap, etc
    (14) pull(obj): pull some object such as rod, handle, etc.
    (15) pour(obj): pour some object using the object in the robot's hand.
    (16) lean(obj1, obj2): lean some object against some other object, such as lean a board against a wall.

    Instructions: {instructions}
    When you make decisions, you should consider constraints based on your common sense. For example, the robot cannot pick up another object when it is already holding one because it is single-armed.
    Structure the code with reusable high-level functions that encapsulate meaningful sub-tasks using atomic action APIs.
    This code will be directly executed, so it must be syntactically correct Python. No markdown formatting like python or text outside the code.
    
    Example: Open the drawer, take out a towel and a tissue, fold both, use the tissue to wipe the table, and place the folded towel neatly on a shelf
    import numpy as np
    from robot_utils import pick_up, put_down, open, close, fold, wipe
    def retrieve_items_from_drawer(drawer, towel, tissue):
        open(drawer)
        pick_up(towel)
        put_down("table")
        pick_up(tissue)
        put_down("table")
    def prepare_items(towel, tissue):
        pick_up(towel)
        fold(towel)
        put_down("table")
        pick_up(towel)
        fold(tissue)
        put_down("table")
    def wipe_table(tissue, table):
        pick_up(tissue)
        wipe(table)
        put_down(tissue)
    def store_towel_on_shelf(towel, shelf):
        pick_up(towel)
        put_down(shelf)
    # Main execution
    retrieve_items_from_drawer("drawer", "towel", "tissue")
    prepare_items("towel", "tissue")
    wipe_table("tissue", "table")
    store_towel_on_shelf("towel", "shelf")
\end{lstlisting}

\newpage
\subsection{Prompt for ReAct.}
\begin{lstlisting}[basicstyle=\ttfamily\small, breaklines=true, breakindent=0pt]
Now you need to help a single-armed robot to plan to finish the given task. The given image shows the initial scene. The available action APIs for the robot are: 
(1) pick up: pick up some object. 
(2) put down: put down the object in the robot's hand on/in some object. 
(3) open: open some object such as drawer, door, etc. 
(4) close: close some object such as drawer, door, etc. 
(5) fold: fold some object such as towel, tissue, etc. 
(6) wipe: wipe some object using the object in the robot's hand. 
(7) scrunch: scrunch some object such as tissue, etc. 
(8) stir: stir some object using the object in the robot's hand. 
(9) push: push some object such as block, etc. (10) slide: slide some object such as block, etc.
(11) press: press some object such as button, etc. 
(12) turn on: turn on some object such as light, tap, etc. 
(13) turn off: turn off some object such as light, tap, etc 
(14) pull: pull some object such as rod, handle, etc. 
(15) pour: pour some object using the object in the robot's hand.
(16) lean: lean some object against some other object, such as lean a board against a wall. 
When you make decisions, you should consider constraints based on your common sense. For example, the robot cannot pick up another object when it is already holding one because it is single-armed.
You MUST use only the exact API formats listed above for your actions, and output the sequence of actions in natural language to complete the tasks (e.g., pick up the green block from table). If you need to think through a problem first, start your response with 'think:' followed by your reasoning, and you'll receive 'OK.' Then provide your action in the proper format.,
    
react_example_1:
Instruction: Put the blue cup into the open drawer.
Initial scene: [Image showing a table with a blue cup, a closed drawer, and other objects]
> think: I need to open the drawer first, then pick up the blue cup, and finally put it in the drawer. The robot has only one arm, so I need to follow this sequence carefully.
OK.
> open the drawer
Image: [Image showing the drawer now open]
Feedback: Action successfully executed. pick up the blue cup from the table.
Image: [Image showing the robot arm holding the blue cup]
Feedback: Action successfully executed.
> put down the blue cup into the drawer
Image: [Image showing the blue cup placed inside the open drawer]
Feedback: Action successfully executed.,

react_example_0: 
Instruction: Fold the towel and place it on the shelf.
Initial scene: [Image showing a bathroom with an unfolded towel on a rack and an empty shelf]
> pick up the towel from the rack
Image: [Image showing the robot arm holding the unfolded towel]
Feedback: Action successfully executed.
fold the towel
Image: [Image showing the robot arm holding a now folded towel]
Feedback: Action successfully executed.
> put down the towel on the shelf
Image: [Image showing the folded towel placed on the shelf]
Feedback: Action successfully executed.
\end{lstlisting}

\newpage
\subsection{Prompt for VLM-PDDL.}
\begin{lstlisting}[basicstyle=\ttfamily\small, breaklines=true, breakindent=0pt]
Now a single-armed robot is going to finish the given task.
The given image shows the initial scene.

The available action for the robot are:
(1) pick up: pick up some object.
(2) put down: put down the object in the robot's hand on/in some object.
(3) open: open some object such as drawer, door, etc.
(4) close: close some object such as drawer, door, etc.
(5) fold: fold some object such as towel, tissue, etc.
(6) wipe: wipe some object using the object in the robot's hand.
(7) scrunch: scrunch some object such as tissue, etc
(8) stir: stir some object using the object in the robot's hand.
(9) push: push some object such as block, etc.
(10) slide: slide some object such as block, etc.
(11) press: press some object such as button, etc.
(12) turn on: turn on some object such as light, tap, etc.
(13) turn off: turn off some object such as light, tap, etc
(14) pull: pull some object such as rod, handle, etc.
(15) pour: pour some object using the object in the robot's hand.
(16) lean: lean some object against some other object, such as lean a board against a wall.

The robot must consider constraints based on common sense when making decisions.
For example, the robot cannot pick up another object when it is already holding one because it is single-armed.

Now you need to help the robot to generate both PDDL domain file and PDDL problem file to finish the task, according to the given image scene and instructions.
Instructions: {instructions}

Your output should include three parts:
(1) reasoning: analyze the image and output the reasons.
(2) domain: output the PDDL domain file.
(3) problem: output the PDDL problem file.

Your output should be in JSON format like below (Do NOT output comments):
{{
    "reasoning:": "your analysis",
    "domain": "your PDDL domain file",
    "problem": "your PDDL problem file"
}}
\end{lstlisting}

\newpage
\subsection{Prompts for ISR-LLM.}  
\textbf{Translator.}
\begin{lstlisting}[basicstyle=\ttfamily\small, breaklines=true, breakindent=0pt]
You are a helpful, pattern-following assistant that translates given task description into Planning Domain Definition Language (PDDL) domain and problem files.
Now you need to help a single-armed robot to plan to finish the given task.
The given image will show the initial scene.

The available action APIs for the robot are:
(1) pick up: pick up some object.
(2) put down: put down the object in the robot's hand on/in some object.
(3) open: open some object such as drawer, door, etc.
(4) close: close some object such as drawer, door, etc.
(5) fold: fold some object such as towel, tissue, etc.
(6) wipe: wipe some object using the object in the robot's hand.
(7) scrunch: scrunch some object such as tissue, etc
(8) stir: stir some object using the object in the robot's hand.
(9) push: push some object such as block, etc.
(10) slide: slide some object such as block, etc.
(11) press: press some object such as button, etc.
(12) turn on: turn on some object such as light, tap, etc.
(13) turn off: turn off some object such as light, tap, etc
(14) pull: pull some object such as rod, handle, etc.
(15) pour: pour some object using the object in the robot's hand.
(16) lean: lean some object against some other object, such as lean a board against a wall.

When you make decisions, you should consider constraints based on your common sense.
For example, the robot cannot pick up another object when it is already holding one because it is single-armed.
Below are some examples of PDDL files for the blocksworld problem.
{examples}
Now instruction is {instruction}.
\end{lstlisting}

\textbf{Planner.}
\begin{lstlisting}[basicstyle=\ttfamily\small, breaklines=true, breakindent=0pt]
You are a confident and pattern-following assistant that determines action sequences to complete a given task, which is described by Planning Domain Definition Language (PDDL) domain and problem files. 

Below are some examples:
{examples}

Now the planning problem is {planning_problem}.
\end{lstlisting}

\textbf{Validator.}
\begin{lstlisting}[basicstyle=\ttfamily\small, breaklines=true, breakindent=0pt]
You are a helpful, pattern-following assistant that examines the correctness of each action during a task planning process. You work like a state machine.

Below are some examples:
{examples}

Question:
initial state: {pddl_init_state}
Goal state: {pddl_goal_state}
Examined action sequence: {action_description}
\end{lstlisting}

\subsection{Prompts for BoN-iVML.}
\textbf{initialization.}
\begin{lstlisting}[basicstyle=\ttfamily\small, breaklines=true, breakindent=0pt]
Now a single-armed robot is going to finish the given task.
The given image shows the initial scene.

Instructions: {instructions}

The available action for the robot are:
(1) pick up: pick up some object.
(2) put down: put down the object in the robot's hand on/in some object.
(3) open: open some object such as drawer, door, etc.
(4) close: close some object such as drawer, door, etc.
(5) fold: fold some object such as towel, tissue, etc.
(6) wipe: wipe some object using the object in the robot's hand.
(7) scrunch: scrunch some object such as tissue, etc
(8) stir: stir some object using the object in the robot's hand.
(9) push: push some object such as block, etc.
(10) slide: slide some object such as block, etc.
(11) press: press some object such as button, etc.
(12) turn on: turn on some object such as light, tap, etc.
(13) turn off: turn off some object such as light, tap, etc
(14) pull: pull some object such as rod, handle, etc.
(15) pour: pour some object using the object in the robot's hand.
(16) lean: lean some object against some other object, such as lean a board against a wall.

The robot must consider constraints based on common sense when making decisions.
For example, the robot cannot pick up another object when it is already holding one because it is single-armed.

Now you are given the natural language instruction. Your task is to generate PDDL domain code. according to the given image scene and instructions.
This includes defining predicates and actions based on the information provided.
Note that individual conditions in preconditions and effects should be listed separately. For example, "object1 is washed and heated" should be considered as two separate conditions: "object1 is washed" and "object1 is heated".

Also, in PDDL, two predicates cannot have the same name even if they have different parameters. 
Each predicate in PDDL must have a unique name, and its parameters must be explicitly defined in the predicate definition. 
It is recommended to define predicate names in an intuitive and readable way. 
Remember: Ignore the information that you think is not helpful for the planning task.

You are only responsible for domain generation. Before you generate the concrete domain code, you should first generate a natural
Language thought about the meaning of each variable, and the step-by-step explanation of the domain code. Even if I didn't provide the exact name of the predicates and actions, you should generate them based on the information provided in the natural language description.

Note that you ONLY need to use PDDL 1.0!

Your output should be in JSON format like below:
{{
    "reasoning": "Analysis the image and output your reasoning, like: predicate_1: the name of predicate_1, explanation of predictate_1, ..., predicate_n: the name of predicate_n, explanation of predictate_n, action_1: the name of action_1, explanation of action, ..., action_n: the name of actio_n, explanation of action_n.",
    "domain": "The concrete pddl code for domain.pddl, in PDDL format."
}}

Now it's your time to generate the solution, you have to follow the format I provided above.
\end{lstlisting}

\textbf{Best-of-N.}
\begin{lstlisting}[basicstyle=\ttfamily\small, breaklines=true, breakindent=0pt]
Now a single-armed robot is going to finish the given task.
The given image shows the initial scene.

Instructions: {instructions}

The available action for the robot are:
(1) pick up: pick up some object.
(2) put down: put down the object in the robot's hand on/in some object.
(3) open: open some object such as drawer, door, etc.
(4) close: close some object such as drawer, door, etc.
(5) fold: fold some object such as towel, tissue, etc.
(6) wipe: wipe some object using the object in the robot's hand.
(7) scrunch: scrunch some object such as tissue, etc
(8) stir: stir some object using the object in the robot's hand.
(9) push: push some object such as block, etc.
(10) slide: slide some object such as block, etc.
(11) press: press some object such as button, etc.
(12) turn on: turn on some object such as light, tap, etc.
(13) turn off: turn off some object such as light, tap, etc
(14) pull: pull some object such as rod, handle, etc.
(15) pour: pour some object using the object in the robot's hand.
(16) lean: lean some object against some other object, such as lean a board against a wall.

The robot must consider constraints based on common sense when making decisions.
For example, the robot cannot pick up another object when it is already holding one because it is single-armed.

Now a sequence of initial PDDL domains is provided, you must choose the best one that is the most relevant to the task according to the given image.

PDDL domains: {domains}

Your output should be in JSON format like below:
{{
    "reasoning": "Analysis the image and output your reasoning.",
    "domain_index": "1"
}}
\end{lstlisting}

\newpage
\textbf{$f_{\mathrm{opt}}\left( \cdot \right) $}
\begin{lstlisting}[basicstyle=\ttfamily\small, breaklines=true, breakindent=0pt]
Now a single-armed robot is going to finish the given task.
The given image shows the initial scene.

Instructions: {instructions}

The available action for the robot are:
(1) pick up: pick up some object.
(2) put down: put down the object in the robot's hand on/in some object.
(3) open: open some object such as drawer, door, etc.
(4) close: close some object such as drawer, door, etc.
(5) fold: fold some object such as towel, tissue, etc.
(6) wipe: wipe some object using the object in the robot's hand.
(7) scrunch: scrunch some object such as tissue, etc
(8) stir: stir some object using the object in the robot's hand.
(9) push: push some object such as block, etc.
(10) slide: slide some object such as block, etc.
(11) press: press some object such as button, etc.
(12) turn on: turn on some object such as light, tap, etc.
(13) turn off: turn off some object such as light, tap, etc
(14) pull: pull some object such as rod, handle, etc.
(15) pour: pour some object using the object in the robot's hand.
(16) lean: lean some object against some other object, such as lean a board against a wall.

The robot must consider constraints based on common sense when making decisions.
For example, the robot cannot pick up another object when it is already holding one because it is single-armed.

Now a PDDL domain about the task based on the image is given with intermediate thoughts explaining each predicate and action 
Your task is to generate critical feedback on the PDDL domain code based on the task and image scene. 
You should evaluate the grammar and logic of the PDDL domain codes, and the logic error in the intermediate thoughts.

natural language chain of thoughts: {thought}
Generated PDDL domain: {domain}

Note that you ONLY need to use PDDL 1.0!

Your output should be in JSON format like below:
{{
    "reasoning": "Analysis the image and output your reasoning.",
    "feedback": "Your final feedback."
}}
\end{lstlisting}

\newpage
\textbf{$f_{\mathrm{update}}\left( \cdot \right) $ }
\begin{lstlisting}[basicstyle=\ttfamily\small, breaklines=true, breakindent=0pt]
Now a single-armed robot is going to finish the given task.
The given image shows the initial scene.

Instructions: {instructions}

The available action for the robot are:
(1) pick up: pick up some object.
(2) put down: put down the object in the robot's hand on/in some object.
(3) open: open some object such as drawer, door, etc.
(4) close: close some object such as drawer, door, etc.
(5) fold: fold some object such as towel, tissue, etc.
(6) wipe: wipe some object using the object in the robot's hand.
(7) scrunch: scrunch some object such as tissue, etc
(8) stir: stir some object using the object in the robot's hand.
(9) push: push some object such as block, etc.
(10) slide: slide some object such as block, etc.
(11) press: press some object such as button, etc.
(12) turn on: turn on some object such as light, tap, etc.
(13) turn off: turn off some object such as light, tap, etc
(14) pull: pull some object such as rod, handle, etc.
(15) pour: pour some object using the object in the robot's hand.
(16) lean: lean some object against some other object, such as lean a board against a wall.

The robot must consider constraints based on common sense when making decisions.
For example, the robot cannot pick up another object when it is already holding one because it is single-armed.

You will be provided a PDDL domain code about the task based on the image and critical feedback on the PDDL domain code based on the task and image. 
Your task is to generate a new PDDL domain code that is more consistent with the task and update the chain of thoughts.

Natural language chain of thoughts at the previous turn: {thought}
Generated PDDL domain at the previous turn: {domain}
The error of the PDDL domain {feedback}

Note that you ONLY need to use PDDL 1.0!

Your output should be in JSON format like below:
{{
    "reasoning": "Analysis the image and output your reasoning.",
    "thought": "your updated thought",
    "domain": "your updated PDDL domain code in PDDL format"
}}
\end{lstlisting}

\newpage
\textbf{Problem Formulation.}
\begin{lstlisting}[basicstyle=\ttfamily\small, breaklines=true, breakindent=0pt]
Now a single-armed robot is going to finish the given task.
The robot's hand is free initially even though it is not shown in the image.
The given image shows the initial scene.

Now you need to help the robot to generate the PDDL problem file to finish the task, according to the given image scene and instructions.
Instructions: {instructions}
Given PDDL domain: {domain}

Note that you ONLY need to use PDDL 1.0!

Your output should be in JSON format like below (Do NOT output comments):
{{
    "reasoning:": "analysis the image and output your analysis",
    "problem": "your PDDL problem file in PDDL format"
}}
\end{lstlisting}

\end{document}